\documentclass[11pt, a4paper, logo, twocolumn, copyright]{gensyn}
\usepackage{microtype}
\usepackage[most]{tcolorbox}

\usepackage{graphicx}
\usepackage{subcaption}
\usepackage{hyperref}
\usepackage{url}
\usepackage{float}
\usepackage{multirow}
\usepackage{booktabs}
\usepackage{wrapfig2}

\usepackage{amsmath}
\usepackage{amssymb}
\usepackage{mathtools}
\usepackage{amsthm}

\theoremstyle{plain}

\theoremstyle{definition}

\theoremstyle{remark}

\usepackage[textsize=tiny]{todonotes}

\title{F-TIS: Harnessing Diverse Models in Collaborative GRPO}

\author[1,2]{Nikolay Blagoev}
\author[1]{O\u{g}uzhan Ersoy}
\author[3]{Wendelin Boehmer}
\author[2,3]{Lydia Yiyu Chen}

\affil[1]{Gensyn}
\affil[2]{University of Neuchatel}
\affil[3]{TU Delft}


\begin{abstract}
Reinforcement learning methods such as GRPO have seen great popularity in LLM post-training. In GRPO, models produce completions to a set of prompts, which are rewarded, and the policy is updated towards the relatively high reward completions. Due to the auto-regressive nature of models, the generation phase of such style of training can be extremely time consuming. As a solution, prior work has sought to distribute the inference step across many nodes, working parallel. These works assume primarily homogeneous models in the training in order to keep samples as close to on-policy as possible. This assumption may be impractical in decentralized systems, where parties with various computes and preferences may wish to collaborate on the same task. Thus, decentralized training requires an approach that can handle heterogeneous models - different models collaborating on the same tasks. However, this leads to highly off-policy samples presented during training, which prior work has identified that off-policy samples can hurt GRPO convergence.
To enable heterogeneity, we propose Filtered Truncated Importance Sampling (F-TIS) - a GRPO-style training paradigm that can use off-policy samples to improve local model's learning. Our framework allows various models to collaborate in the same RL training run while being communication efficient. We extensively evaluate F-TIS in various heterogeneous setups and we show that it exhibits identical final model convergence to purely on-sample training. Furthermore, we observe in some setups better generalization on out-of-distribution tasks than on-policy training, increasing model's performance by up to 12\%.
\end{abstract}

\begin{document}

\maketitle

\section{Introduction}
\par Reinforcement Learning (RL) has seen a great adoption for the post-training of Large Language Models (LLMs) \cite{grpo,dpo,tradingr1}. Algorithms such as Proximal Policy Optimization (PPO) allow LLMs to learn user-preferred behaviour \cite{ppo}, such as adhering to some ethical code during conversations. More recently, RL has been utilized to a great degree for the purposes of improving the reasoning of LLMs. This is in part due to the highly influential work of \cite{grpo}, which proposed Group Relative Policy Optimization (GRPO). GRPO removes the need for a value-model in PPO, thus greatly reducing the computational and memory requirements, by instead relying on the group relative advantage. In GRPO, for each prompt, multiple completions are generated and each completion's advantage is computed relative to the other completions of the same prompt. Several works have demonstrated the success of GRPO-based algorithms in improving an LLM's reasoning for various tasks \cite{tradingr1,drgrpo,deepseekr1}.

\par Despite introducing lower memory footprint relative to PPO, GRPO comes with a high computational cost of generating multiple completions for each group (often 8 or more) \cite{llamarl}. This results in a bottleneck during the generation step, as LLMs generate in an auto-regressive manner, i.e., one token at a time~\cite{infinitesampler}. A natural step is to distribute the generation step across many workers thus speeding up the generation \cite{llamarl,intellect2}. Such systems, however, assume relatively homogeneous models and resources, making them impractical for decentralized training. Furthermore, even if participants start from the same models, if they solely communicate completions, their models will drift apart, due to floating-point non-associativity. Such drifts introduce off-policyness between the generator and the trainer, harming the final model's performance \cite{revisitinggrpo,yao2025efficient_rl_offpolicy}.

\par Beyond these emergent drifts, we identify three types of model heterogeneity in decentralized RL post-training:
\begin{itemize}
    \item \textbf{Model Size}: Models of various sizes (e.g. of 1 billion parameters and 3 billion parameters) can be trained together in the same loop. More constrained devices train the smaller models, while larger devices or clusters can accommodate larger models. Alternatively, one can alleviate the costs of sampling from a large model (as the auto-regressive inference of LLMs is the biggest bottleneck in GRPO training \cite{infinitesampler}). Thus they can utilize some number of small models to populate the batch size. 
    \item \textbf{Model Expertise}: Models of similar size, but different parameters and expertise can be trained together, as different users might have a personalized model they want to collaboratively improve on some task.
    \item \textbf{Trainable Parameters}: Some nodes may train a different subset of the same model's parameters (e.g. with parameter efficient fine-tuning). Such cases can arise due to resource constraints on some devices, needing a smaller set of trainable parameter, or to avoid destructive interference with other tasks \cite{aim}.
\end{itemize}

\par We empirically show that GRPO underperforms in such relative to training purely on-policy. To address this challenge, we propose F-TIS - a single unified framework that can make use of off-policy samples in decentralized heterogeneous Reinforcement Learning. This framework requires communication volume of only \(8\times |p|\) bytes, i.e., 8 bytes per each token, as it only communicates the log-probabilities and tokens of completions between nodes. Despite this minimal overhead, F-TIS can achieve performance identical to homogeneous RL and can even improve model's reasoning capabilities on out-of-distribution tasks (by up to 12\%). Beyond its benefits in decentralized training, F-TIS provides benefits even for a single cluster-training, where multiple models can be post-trained in one training run, rather than several sequential ones.

\section{Related Work}
\paragraph{Reinforcement Learning} has been commonly used for LLM post-training. Proximal Policy Optimization allowed models to learn user preference, which were implicitly hidden in data \cite{ppo}. More recently, GRPO and its derivatives have been utilized to boost model's reasoning and instruction-following abilities. When a model \(\theta\) is trained with GRPO, it generates a number \(G\) of completions (responses) \(a_i\) per prompt \(p\) (\(p\circ a_i\;\forall i \in G\) where \(\circ\) is a concatenation operation), which is termed a "group". Each of the completions is rewarded via some reward model, to get a single scalar value \(r_i\). Commonly used reward mechanisms in GRPO are rule-based rewards~\citep{drgrpo}, which check if the formatting and final answer is correct for a math task or all tests pass for a coding tasks. To replace the need for a value model, GRPO uses the group  advantage \(\hat{A}_i\), \(\hat{A}_i := \frac{r_i - \mu_r}{\sigma_r}\), where \(\mu\) and \(\sigma\) are the mean and standard deviation of the rewards for the completions belonging to the same prompt. 
The advantage is then used to compute the loss:
\begin{align*}
    \mathcal{L}_{GRPO} = \frac{1}{G}\sum_{i=1}^G \frac{1}{|a_i|}\sum_{t=1}^{|a_i|}min[\frac{\pi_{\theta}(a_{i,t}|p\circ a_{i,<t})}{\pi_{\theta_{gen}}(a_{i,t}|p\circ a_{i,<t})}\hat{A}_i,\\
    clip(\frac{\pi_{\theta}(a_{i,t}|p\circ a_{i,<t})}{\pi_{\theta_{gen}}(a_{i,t}|p\circ a_{i,<t})}\hat{A}_i,1-\epsilon, 1+\epsilon)]
\end{align*}
where \(\pi_{\theta_{gen}}\) is the policy that generated the completion. 

Training with GRPO can be divided into two phases - completion generation (gathering as many completions for various prompts) and training (computing the gradient based on the loss for all completions and for all prompts) \citep{llamarl}. Prior work has identified that the KL term (\(\mathcal{D}_{KL}(\pi_{\theta}\parallel\pi_{\theta_{ref}})\), does not benefit the training and increases memory and computational demands \cite{drgrpo}. In line with these works, we omit the KL-term in our experiments.

\paragraph{Distributed RL} has gained popularity as a means of speeding up the generation phase of GRPO-style training \cite{llamarl}. Deployed systems have even successfully trained models in a decentralized manner, however still assuming homogeneous models in order to prevent off-policy issues \cite{genrl,intellect2}. Such approaches have been classified into two categories: \textit{vertical} and \textit{horizontal} \cite{httt}. In vertical RL, each device/node generates the entire group for one prompt. In horizontal RL, each device generates a subset of the group for all prompts (the same across devices). At the end of the generation phase, an all-gather is performed across devices, synchronizing the completions. 

\section{Decentralized Heterogeneous RL}
\par GRPO is fundamentally an on-policy algorithm \cite{grr}. While the clipped importance sampling provides some tolerance for "stale" or off-policy samples, prior work has identified that even small divergence due to difference in probabilities between the inference and training engine can lead to degradation in the RL training \cite{yao2025efficient_rl_offpolicy}. This acts as destructive noise that over time can collapse the policy.
\par
\begin{figure}[tb]
    \centering
    \begin{subfigure}{0.23\textwidth}
        \centering
        \includegraphics[width=\textwidth]{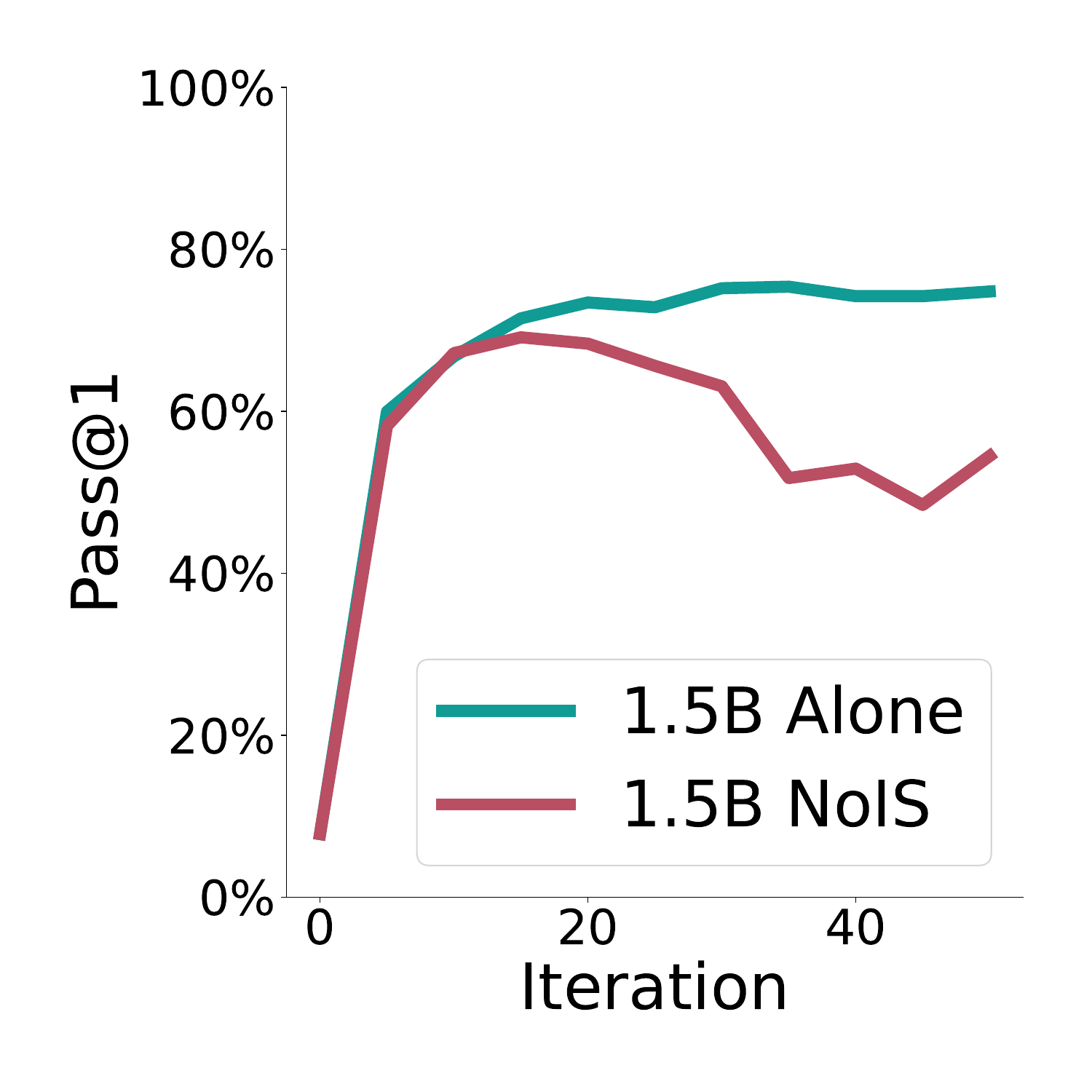}
        
        \caption{1.5B model}
        
    \end{subfigure}
    \begin{subfigure}{0.23\textwidth}
        \centering
        \includegraphics[width=\textwidth]{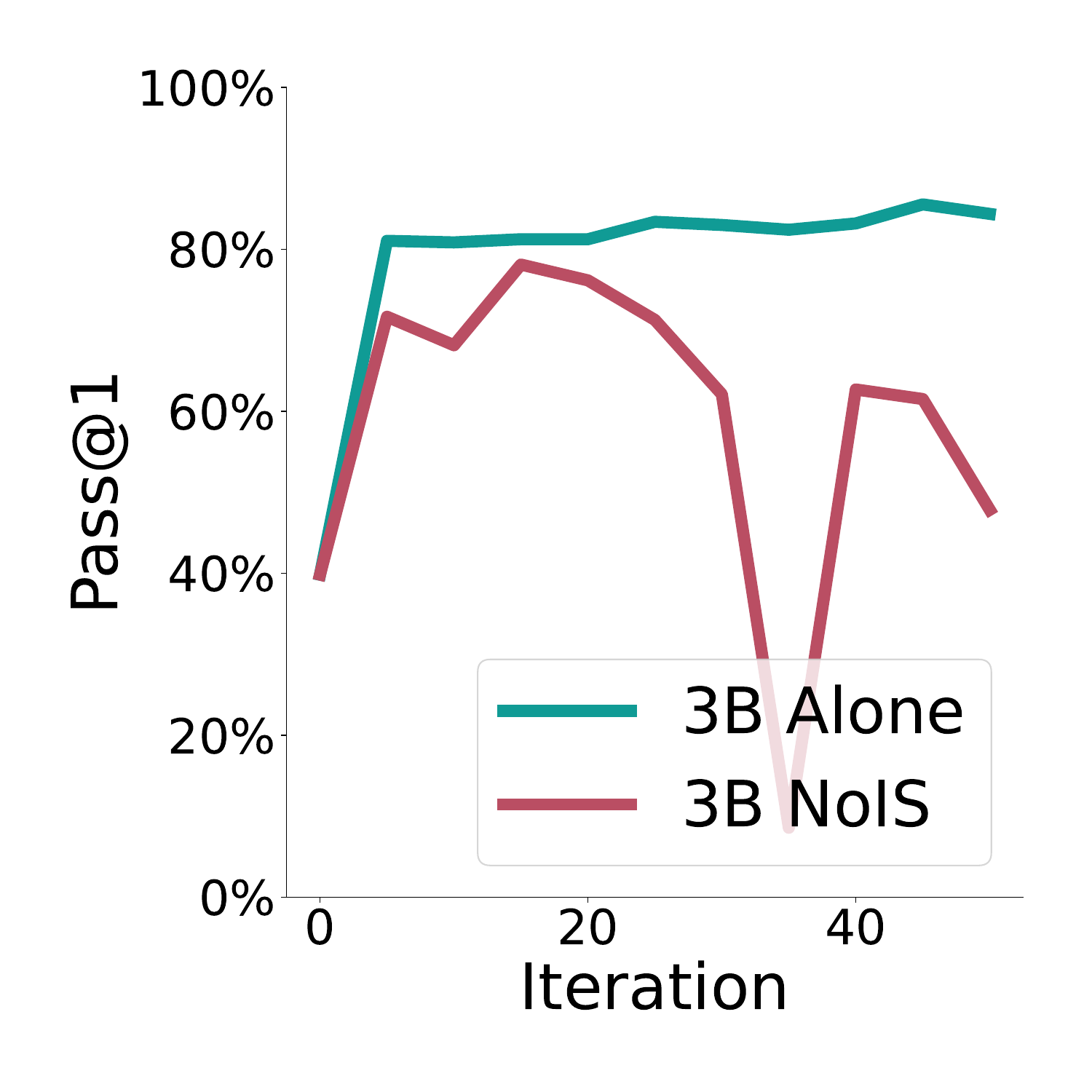}
        
        \caption{3B model}
        
    \end{subfigure}
    \caption{Validation curves of two models collaboratively trained via GRPO. NoIS presents heterogeneous training with no importance sampling used.}
    \label{fig:nis}
\end{figure}

 We replicate these findings for decentralized RL with a motivational example of two models of different sizes (Qwen2.5-1.5B and Qwen2.5-3B \cite{qwen2.5}) train collaboratively via vertical decentralized RL on the GSM8k dataset \cite{cobbe2021gsm8k} (details of all hyperparameters can be found in Table~\ref{tab:hyper}). Here we employ the simplest approach - treat all samples as if on-policy, what we will term NoIS (No Importance Sampling) \cite{genrl}.
We present the validation curves in Figure  \ref{fig:nis}, where we show that both models have significantly worse performance than if they were to train alone, making collaborative training unattractive for different users.

\subsection{Importance Sampling}

Prior work has proposed two modifications to the GRPO loss to account for off-policy examples. The first we term VIS (Vanilla Importance Sampling), which communicates each time the \(\pi_{\theta_{gen}}(a_{i,t}|p\circ a_{i,<t})\) for every generation \cite{llamarl,grpo}. A different approach, proposed by \cite{yao2025efficient_rl_offpolicy}, which they term Truncated Importance Sampling (TIS), takes the importance sampling term outside the inner-most sum's term:
\begin{align*}
    \mathcal{L}_{GRPO} = \frac{1}{G}\sum_{i=1}^G \frac{1}{|a_i|}\sum_{t=1}^{|a_i|}min\left(\frac{\pi_{\theta}(a_{i,t}|p\circ a_{i,<t})}{\pi_{\theta_{gen}}(a_{i,t}|p\circ a_{i,<t})},C\right)\\
    min\left(\mathcal{R}_{i,\theta}\hat{A}_i, clip(\mathcal{R}_{i,\theta}\hat{A}_i,1-\epsilon, 1+\epsilon)\right)
\end{align*}
for some constant \(C\), where \(\mathcal{R}_{i,\theta} = \frac{\pi_{\theta}(a_{i,t}|p\circ a_{i,<t})}{\pi_{\theta_{detach}}(a_{i,t}|p\circ a_{i,<t})}\). 

\par
\begin{figure}[tb]
    \centering
    \begin{subfigure}{0.23\textwidth}
        \centering
        \includegraphics[width=\textwidth]{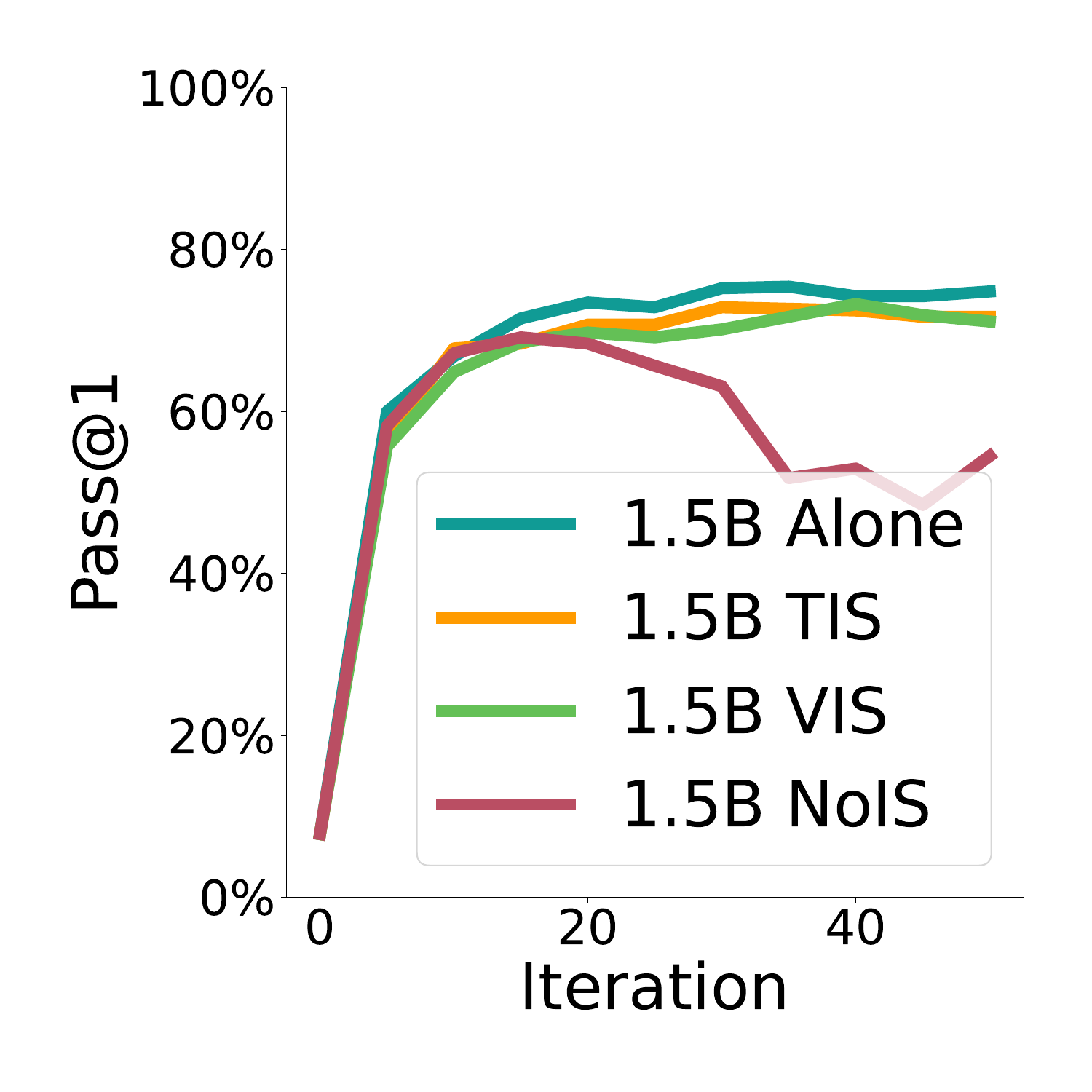}
        
        \caption{1.5B model}

    \end{subfigure}
    \begin{subfigure}{0.23\textwidth}
        \centering
        \includegraphics[width=\textwidth]{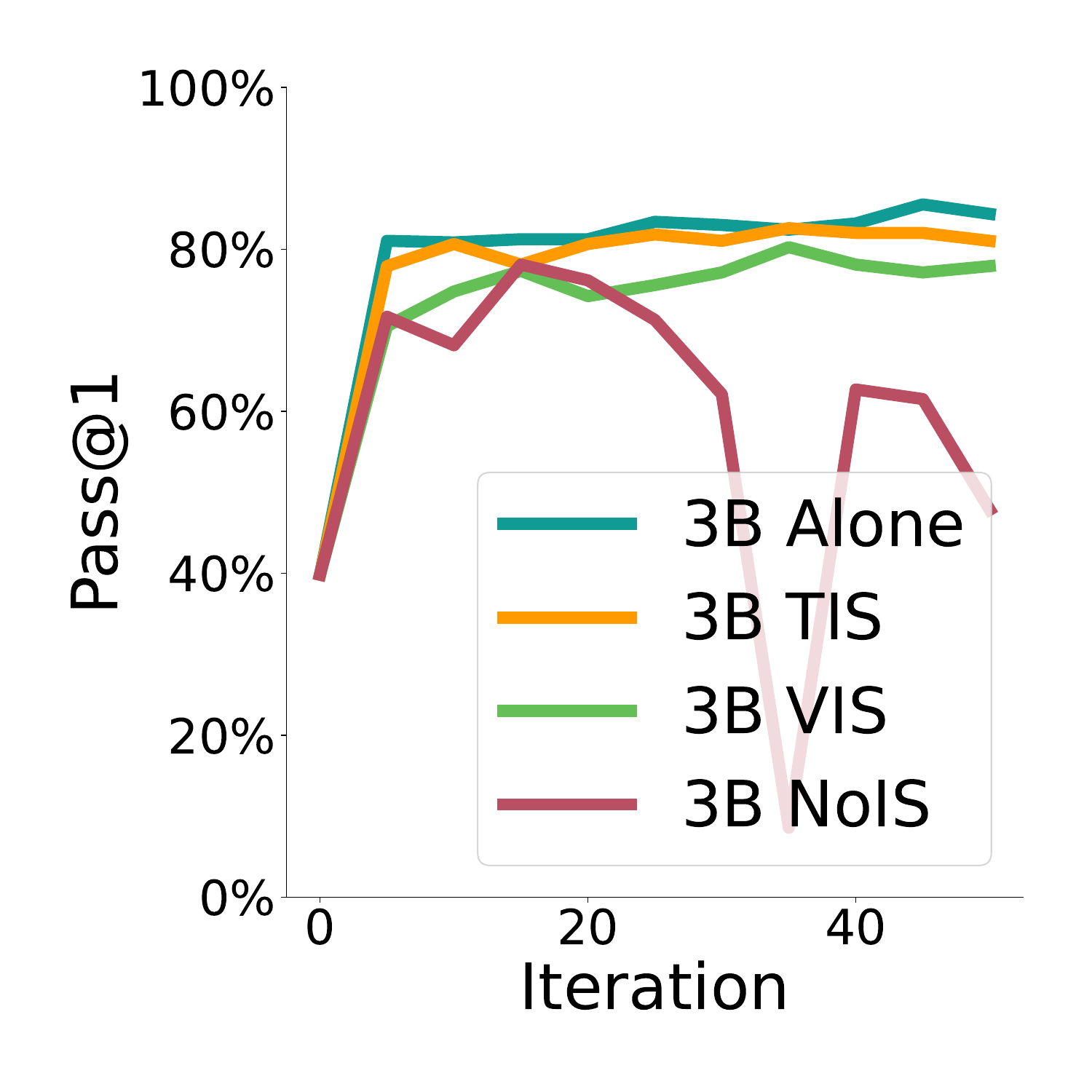}
        
        \caption{3B model}
        
    \end{subfigure}
    \caption{Comparison of various importance sampling methods - NoIS (No Importance Sampling), VIS (Vanilla Importance Sampling), and TIS (Truncated Importance Sampling).}
    \label{fig:nis_vis_tis}
\end{figure}

 We compare the three methods (NoIS, VIS, and TIS) in Figure \ref{fig:nis_vis_tis}, with \(C=2\). While for the smaller model, TIS and VIS behave almost identically, for the larger model TIS is clearly superior than VIS. This is somewhat expected as previous work has suggested that TIS offers better performance than VIS \cite{yao2025efficient_rl_offpolicy}.

\par These approaches require the additional communication only of per-token log-probabilities, which are small in size (4 bytes per token), thus satisfying our requirement for introducing minimal communication overhead.

\subsection{Filtering Samples}
\par
\begin{figure}[tb]
    \centering
    \begin{subfigure}{0.23\textwidth}
        \centering
        \includegraphics[width=\textwidth]{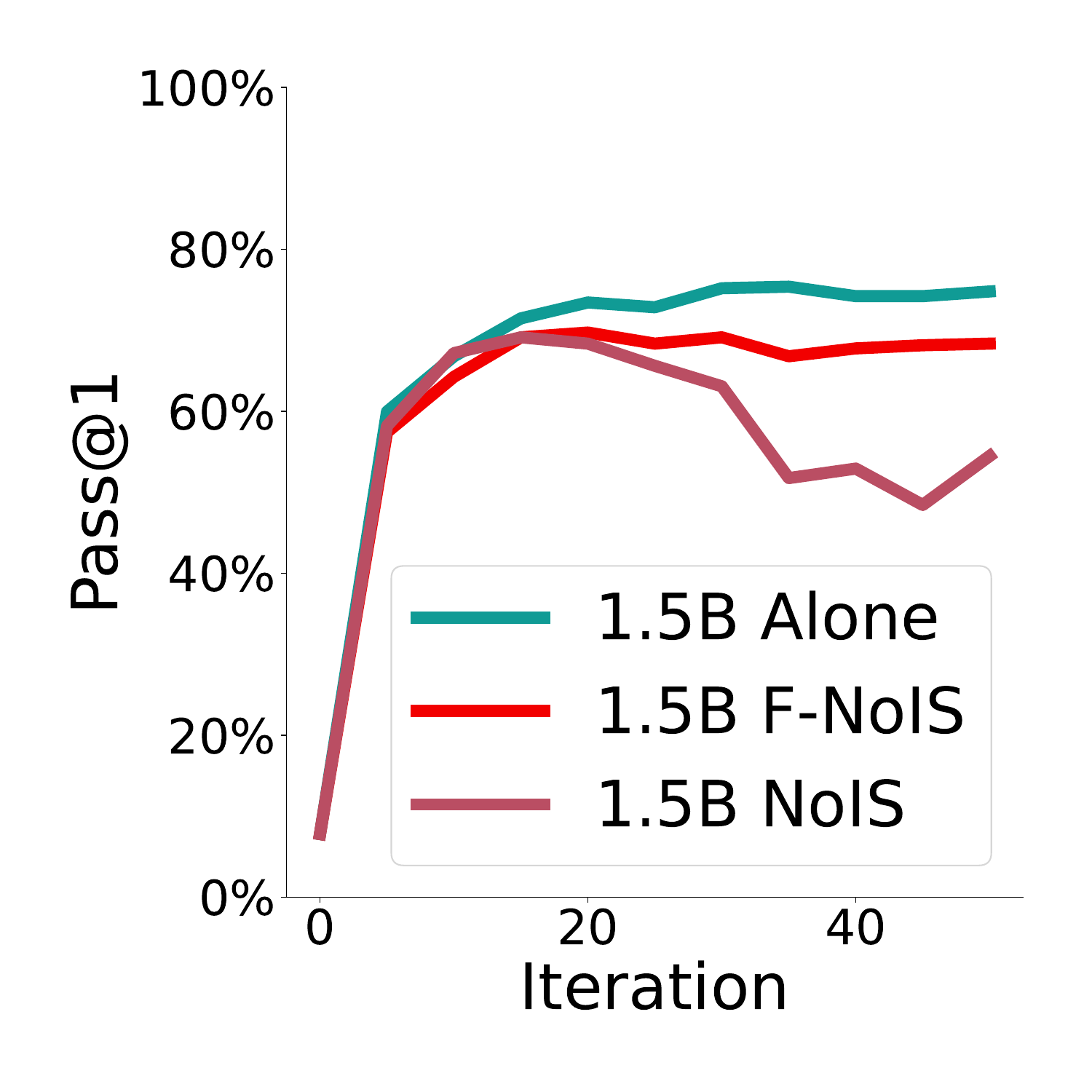}
        
        \caption{1.5B model}

    \end{subfigure}
    \begin{subfigure}{0.23\textwidth}
        \centering
        \includegraphics[width=\textwidth]{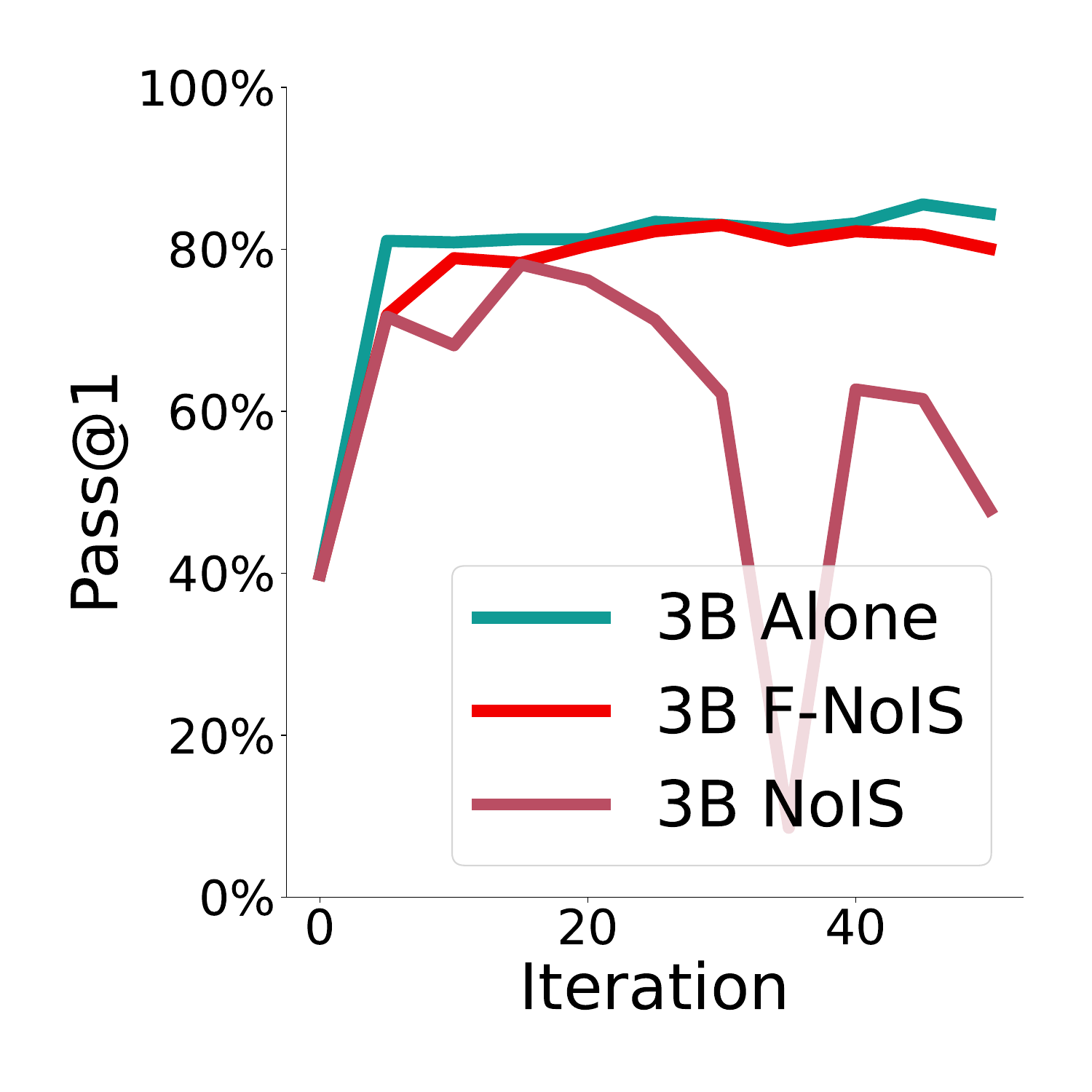}
        
        \caption{3B model}

    \end{subfigure}
    \caption{The effect of filtering in heterogeneous RL. F-NoIS presents a filtered version of NoIS.}
    \label{fig:f_results}
\end{figure}

 Another line of work has explored filtering low-quality off-policy samples as an approach of dealing with off-policy samples \cite{deepseek3,httt}.
During the update phase, samples with advantage less than 0, \(\hat{A}_i<0\), and with KL-divergence beyond some threshold \(g\), \(\mathcal{D}_{KL}(\pi_{\theta}\parallel\pi_{\theta_{gen}}) > g \), are used only to compute the group advantage, but are filtered out before the update phase. The intuition is that samples with \(\hat{A}_i>0\) provide a direction for the policy to move towards, even if off-policy. Off-policy samples with  \(\hat{A}_i<0\) do not - they end up amplifying less probable tokens, which the model would not have produced, resulting often times in gibberish completions. We evaluate a filtered version of NoIS, F-NoIS, with \(g = 50\). The validation curves in Figure \ref{fig:f_results} demonstrate a clear improvement over its non-filtered counterpart, stabilizing the training and preventing model collapse. Even just filtering provides performance close to that of the baseline.

\subsection{F-TIS}

\par Our approach combines the performance of TIS with the stability of filtering into Filtered Truncated Importance Sampling (F-TIS). The modified GRPO formula can thus be written as:
\begin{align*}
    \mathcal{L}_{GRPO} = \frac{1}{G}\sum_{i=1}^G \frac{1}{|a_i|}\sum_{t=1}^{|a_i|}min\left(\frac{\pi_{\theta}(a_{i,t}|p\circ a_{i,<t})}{\pi_{\theta_{gen}}(a_{i,t}|p\circ a_{i,<t})},C\right)\\
    \quad min\left(\mathcal{R}_{i,\theta}\hat{A}_{t,i}, clip(\mathcal{R}_{i,\theta}\hat{A}_{t,i},1-\epsilon, 1+\epsilon)\right)\\
  \hat{A}_{t,i} =
  \begin{cases}
    \hat{A}_i & \text{if $\hat{A}_i > 0$ or $\mathcal{D}_{KL}(\pi_{\theta}\parallel\pi_{\theta_{gen}}) < g$} \\
    0 & \text{else}
  \end{cases}
\end{align*}

\section{Results}
\par Throughout these experiments, unless otherwise specified, we train two models via vertical decentralized RL, with a group size of 12 and batch size of 24. We use \(g=50\) and \(C=2\). All models are trained on the GSM8k \cite{cobbe2021gsm8k} dataset for 50 iterations and evaluated on a held-out validation set via greedy-decoding (\(pass@1\)).

We further test the final models' performance on an out-of-distribution dataset - MATH-500 \cite{huggingfaceh4_math500}. Additional information on the hyperparameters can be found in Table~\ref{tab:hyper}. We use a binary reward function \cite{drgrpo}, which is 1 if and only if the formatting and final answer are correct, with 0 otherwise. Details on the system prompt used can be found below.

\begin{tcolorbox}[colback=blue!5!white,colframe=blue!75!black,title=System prompt for experiments]
\small
  \texttt{A conversation between User and Assistant. The user asks a question, and the Assistant solves it.\\
The assistant needs to provide a detailed step by step solution of the problem. The reasoning process is enclosed within <think> </think> and the answer within  <answer> </answer< tags, i.e.,  <think> reasoning process here  </think>\\
 <answer> answer here </answer>}
\end{tcolorbox}

\begin{table}[b]
    \centering
    \begin{tabular}{c|c}
       Parameter & Value \\
       \hline
        learning rate & \(1\times10^{-6}\)\\
        group size & \(12\) \\
        batch size & \(16\) \\ 
        \(\epsilon\) & \(0.2\) \\
        \(C\) & \(2\) \\
        \(g\) & \(50\) \\
    \end{tabular}
    \caption{Hyperparameters for the experiments.}
    \label{tab:hyper}
\end{table}

\subsection{Model Size Heterogeneity}
\label{sec:results_size}
\par
\begin{figure}[tb]
    \centering
    \begin{subfigure}{0.23\textwidth}
        \centering
        \includegraphics[width=\textwidth]{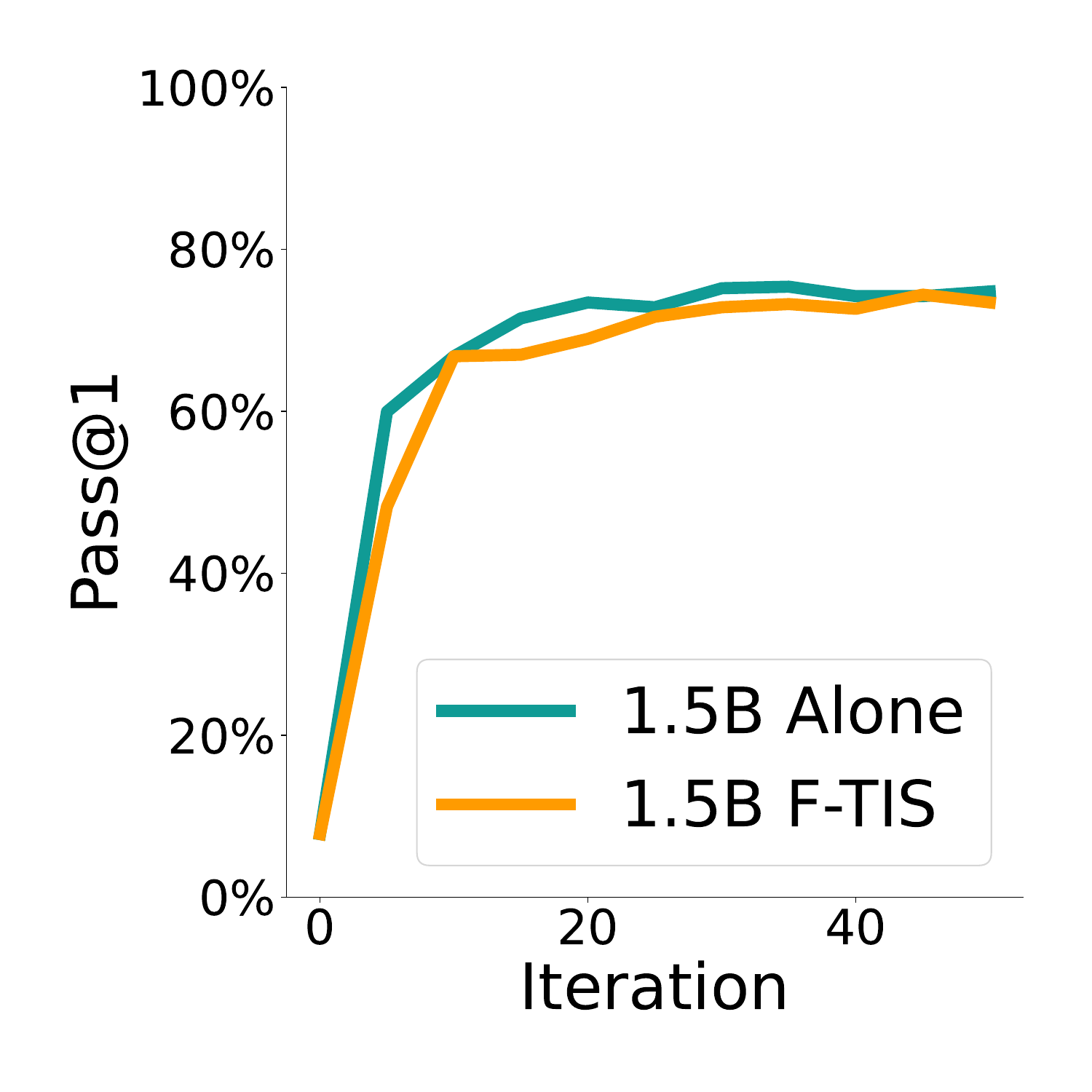}
        
        \caption{1.5B model}

    \end{subfigure}
    \begin{subfigure}{0.23\textwidth}
        \centering
        \includegraphics[width=\textwidth]{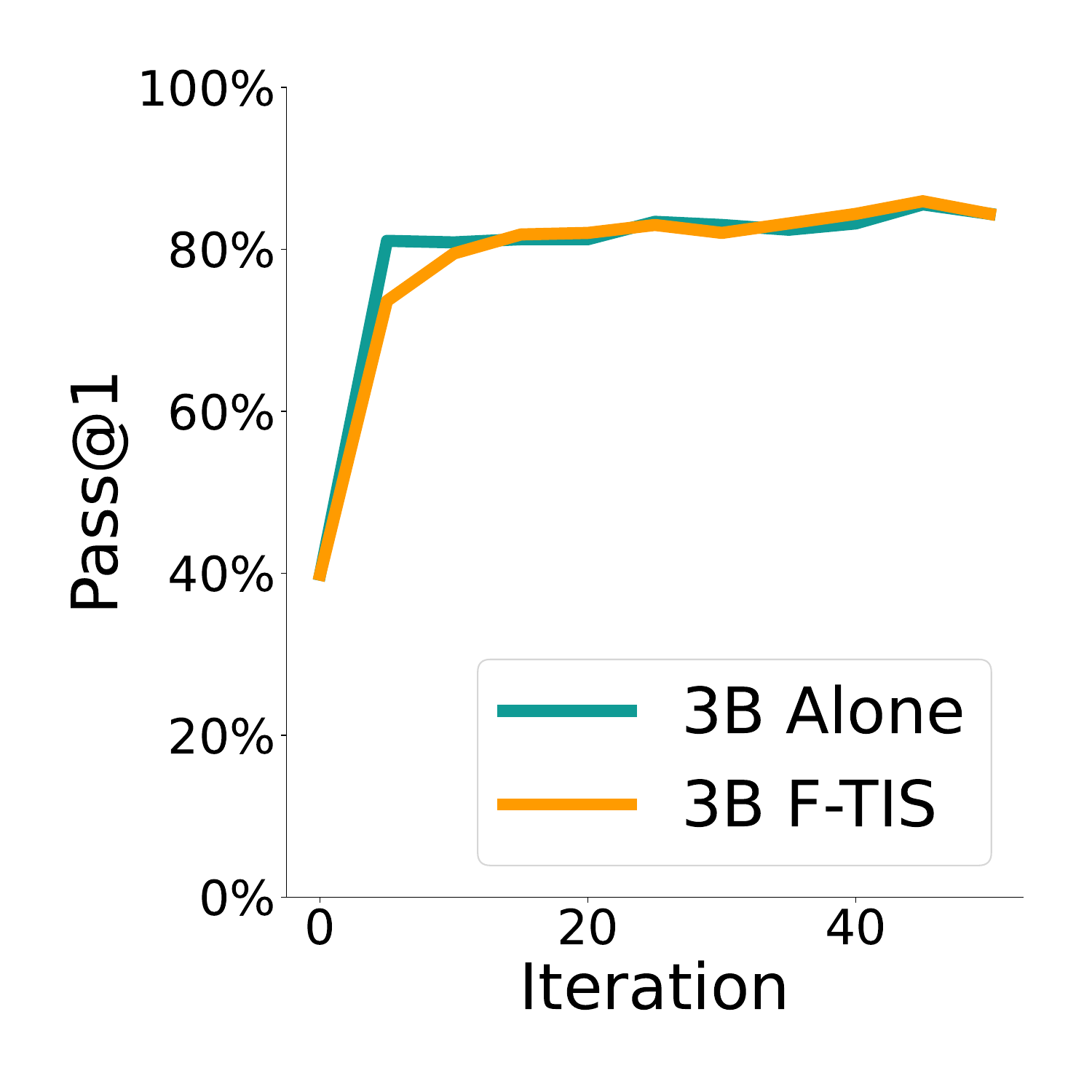}
        
        \caption{3B model}

    \end{subfigure}
    \caption{Validation curves of a Qwen2.5-1.5B and a Qwen2.5-3B trained together.}
    \label{fig:base_size_results}
\end{figure}

\begin{figure}[tb]
    \centering
    \begin{subfigure}{0.23\textwidth}
        \centering
        \includegraphics[width=\textwidth]{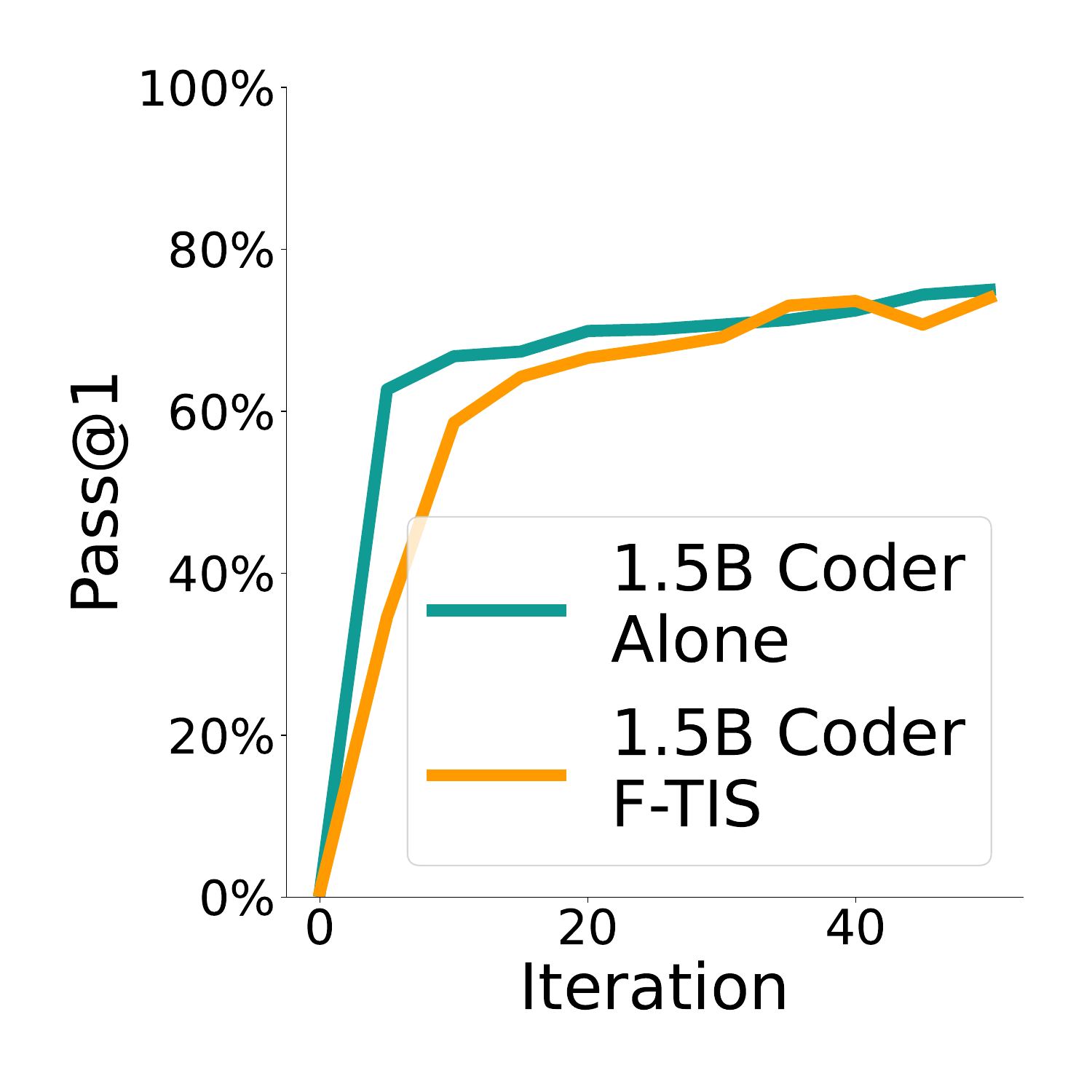}
        
        \caption{1.5B Coder model}

    \end{subfigure}
    \begin{subfigure}{0.23\textwidth}
        \centering
        \includegraphics[width=\textwidth]{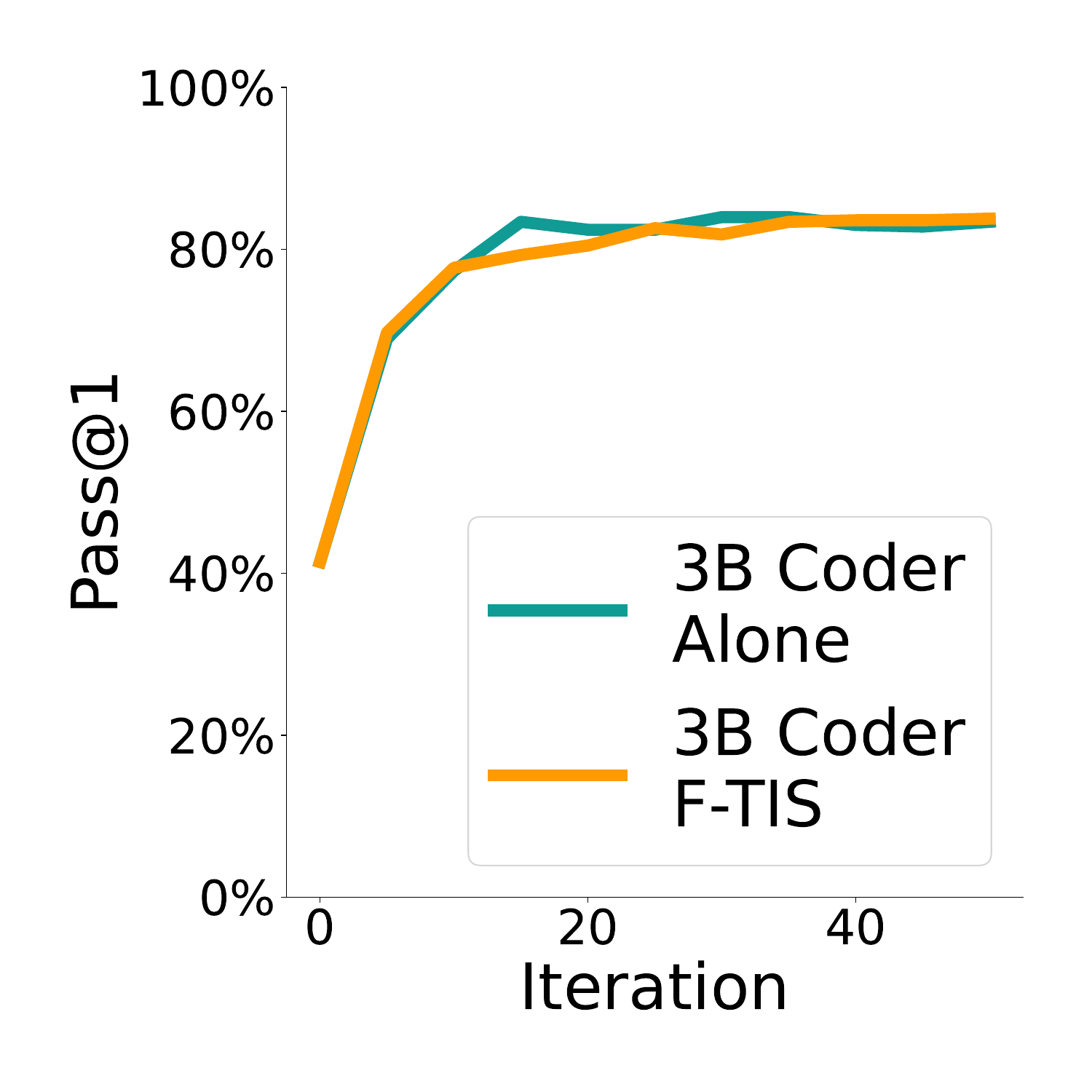}
        
        \caption{3B Coder model}

    \end{subfigure}
    \caption{Validation curves of a Qwen2.5-Coder-1.5B and a Qwen2.5-Coder-3B trained together.}
    \label{fig:coder_size_results}
\end{figure}

 We perform two sets of experiments in vertical decentralized RL with F-TIS: collaborative training of Qwen2.5-1.5B and Qwen2.5-3B models, and collaborative training of Qwen2.5-Coder-1.5B and Qwen2.5-Coder-3B models. The validation curves during training are reported in Figures \ref{fig:base_size_results} and \ref{fig:coder_size_results}. We observe near identical final performance of collaborative training with F-TIS to the baseline. However, across all experiments we observe an initially slower convergence relative to the baseline, which we discuss further in Section \ref{sec:g_ablation}. We report the performance on the MATH-500 dataset of the models in Table \ref{tab:math500}, where we observe that on out-of-distribution tasks the smaller models display a significant improvement in performance.

\begin{table}
    \centering
    \begin{tabular}{|c|c|c|}

    \hline
          & Model & Math-500\\
    \hline
    \multirow{6}{*}{Alone}  & 1.5B Base & 0.406\\
                            & 3B Base & 0.575\\
                            & 1.5B Coder & 0.41 \\
                            & 3B Coder & 0.478 \\
                            & 1.5B PEFT & 0.412 \\
                            & 3B PEFT &  0.5\\
 
    \hline
    \multirow{2}{*}{F-TIS}  & 1.5B Base & 0.47 \\
                            & 3B Base & 0.54 \\

    \hline
    \multirow{2}{*}{F-TIS}  & 1.5B Coder & 0.47 \\
                            & 3B Coder & 0.59 \\ 
   
    \hline
    \multirow{2}{*}{F-TIS}  & 1.5B Base & 0.403 \\
                            & 1.5B Coder & 0.41 \\

    \hline
    \multirow{2}{*}{F-TIS}  & 3B Base & 0.52 \\
                            & 3B Coder & 0.53 \\
    \hline
    \multirow{2}{*}{F-TIS}  & 1.5B Base & 0.43 \\
                            & 1.5B PEFT & 0.43 \\
    \hline
    \multirow{2}{*}{F-TIS}  & 3B Base & 0.513 \\
                            & 3B PEFT & 0.56 \\
    \hline
\end{tabular}
    \caption{Final model evaluation on the Math-500 dataset.}
    \label{tab:math500}
\end{table}

\subsection{Model Expertise Heterogeneity}
\par
\begin{figure}[tb]
    \centering
    \begin{subfigure}{0.23\textwidth}
        \centering
        \includegraphics[width=\textwidth]{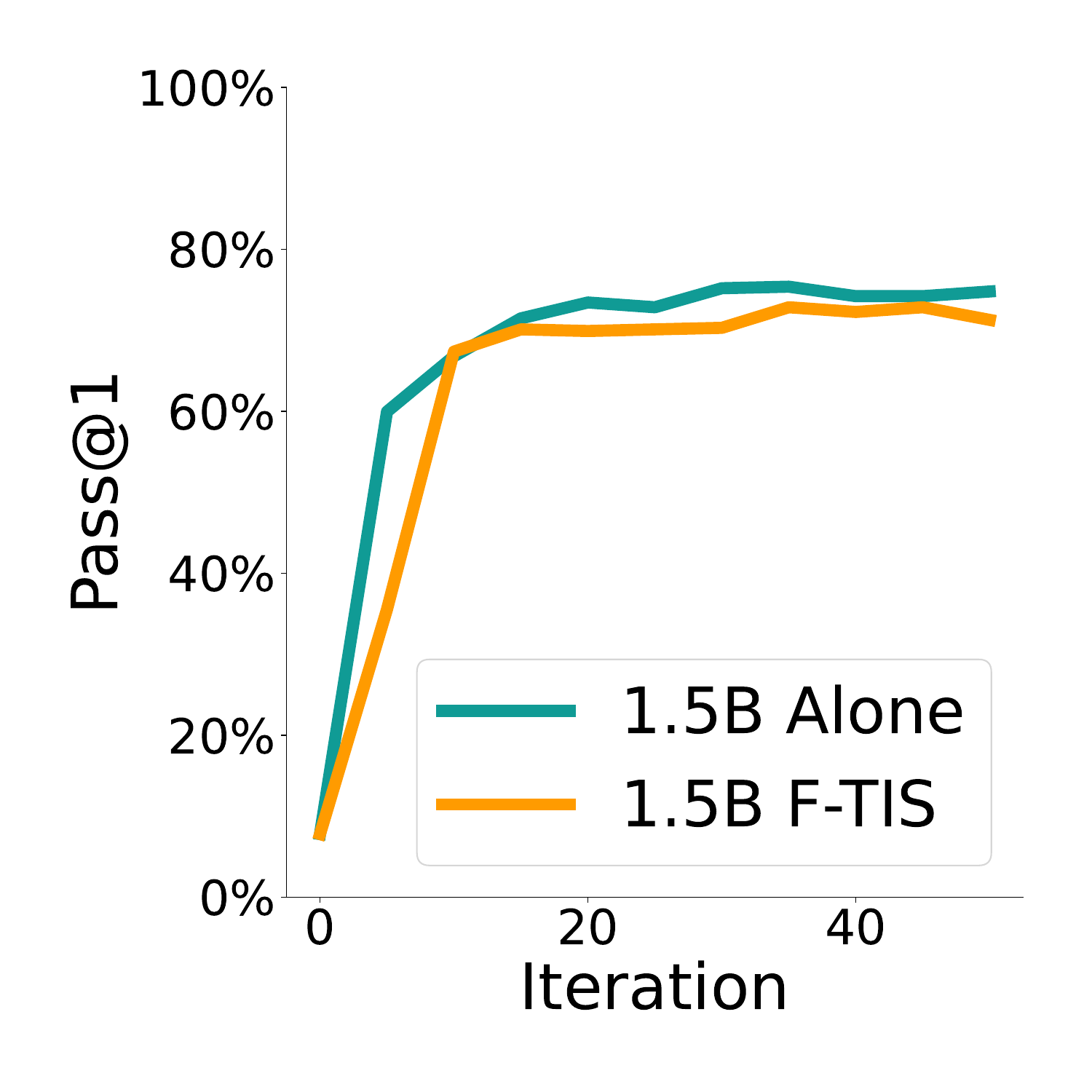}
        
        \caption{1.5B model}

    \end{subfigure}
    \begin{subfigure}{0.23\textwidth}
        \centering
        \includegraphics[width=\textwidth]{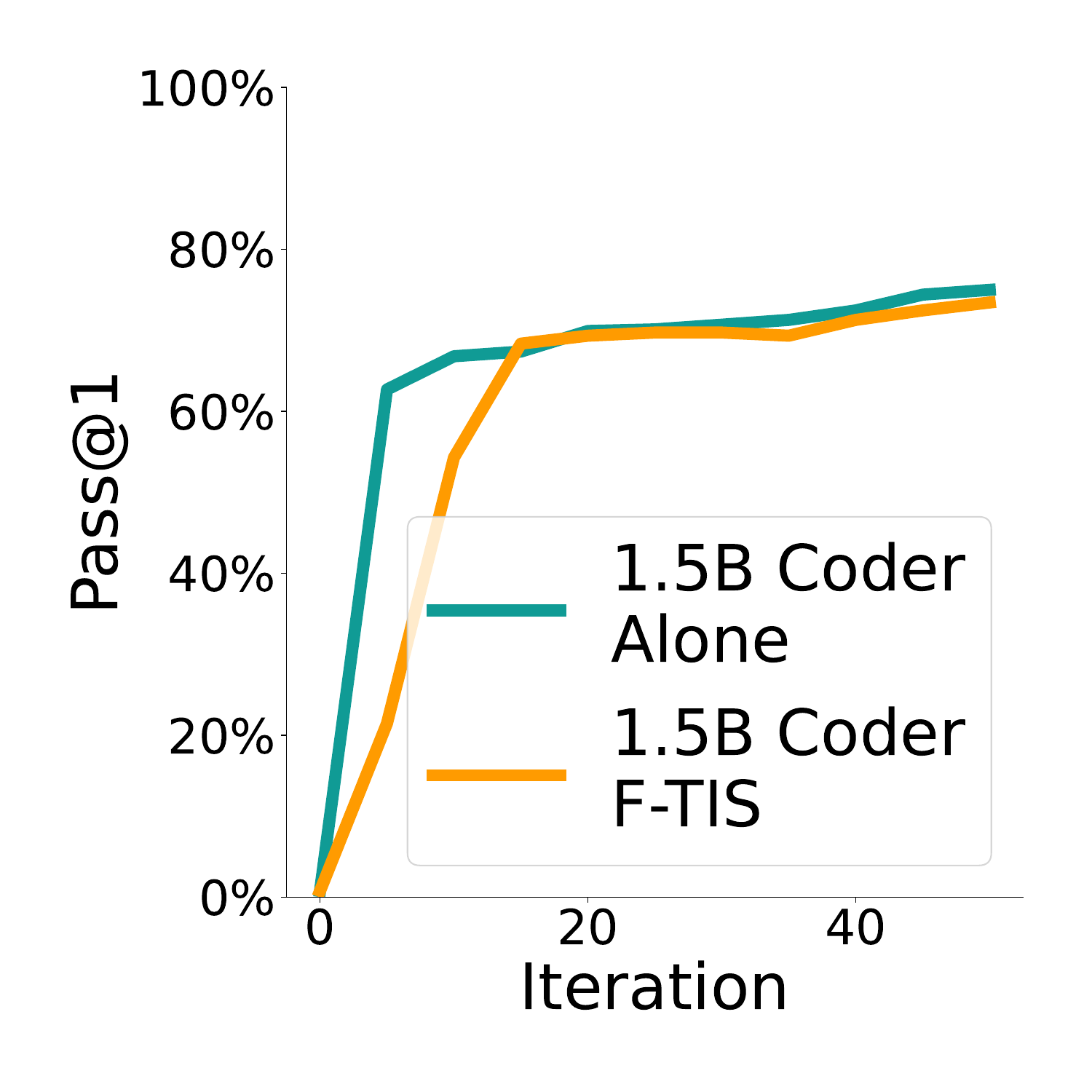}
        
        \caption{1.5B Coder model}

    \end{subfigure}
    \caption{Validation curves of a Qwen2.5-1.5B and a Qwen2.5-Coder-1.5B trained together.}
    \label{fig:1b_exp_results}
\end{figure}

\begin{figure}[tb]
    \centering
    \begin{subfigure}{0.23\textwidth}
        \centering
        \includegraphics[width=\textwidth]{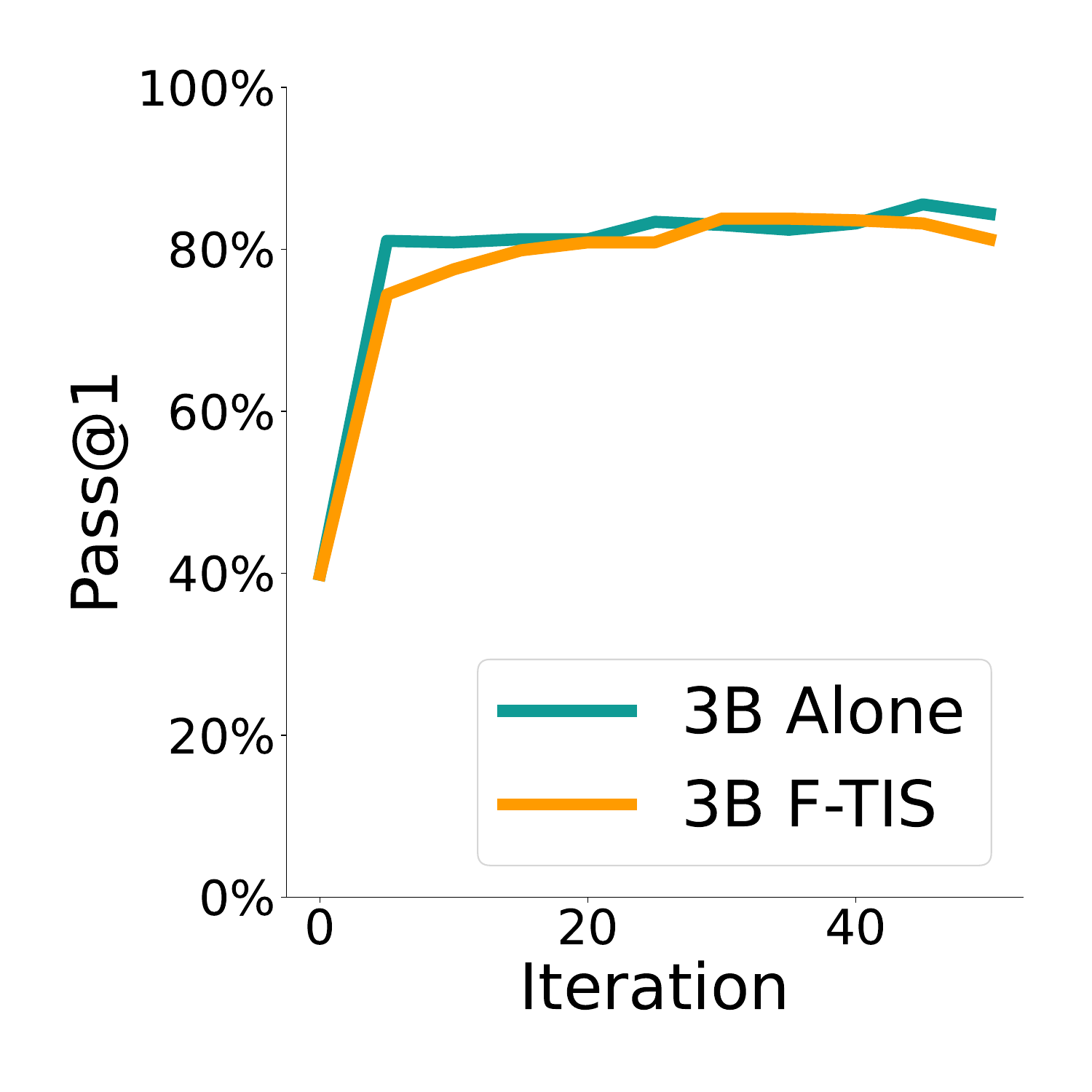}
        
        \caption{3B model}
    \end{subfigure}
    \begin{subfigure}{0.23\textwidth}
        \centering
        \includegraphics[width=\textwidth]{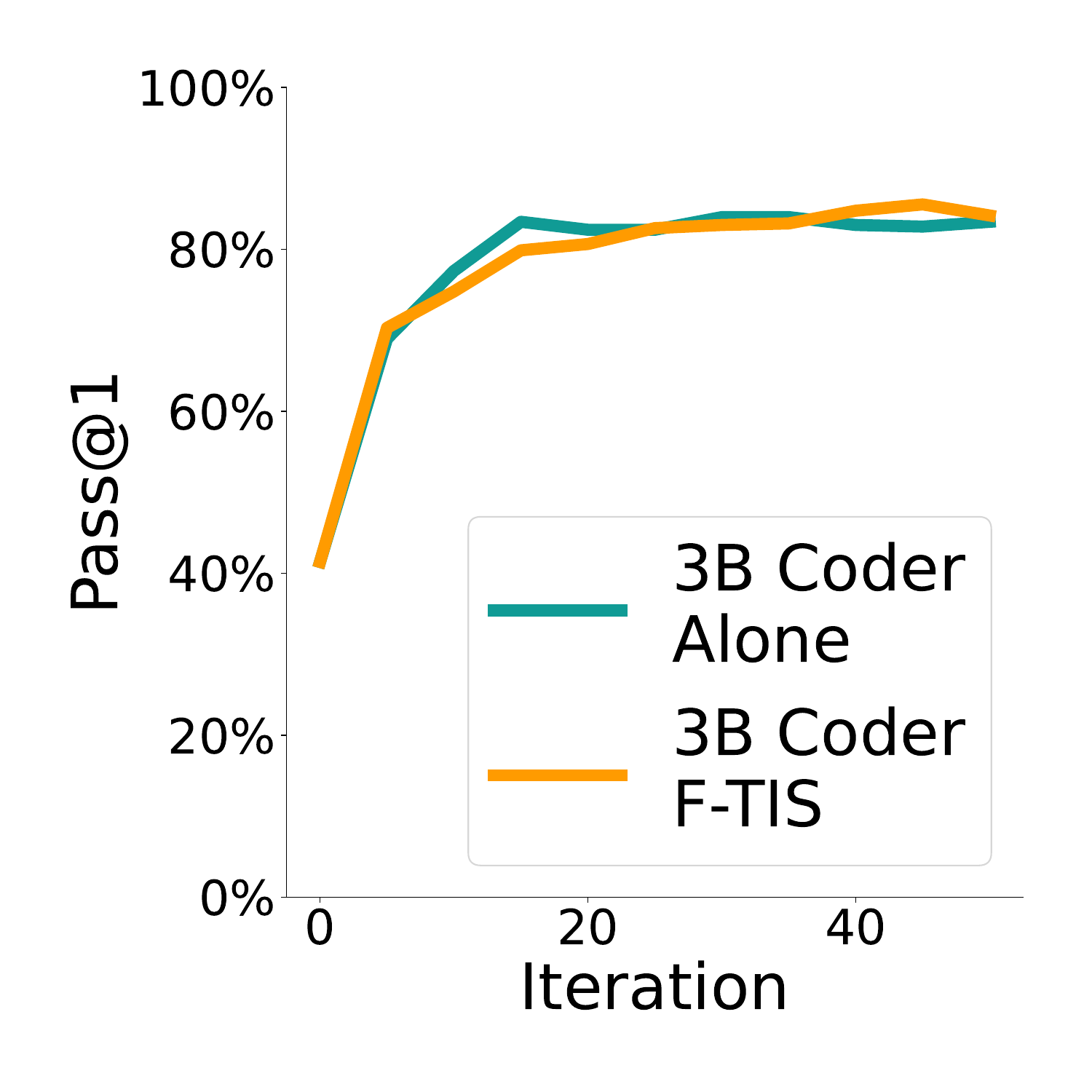}
        
        \caption{3B Coder model}
    \end{subfigure}
    \caption{Validation curves of a Qwen2.5-3B and a Qwen2.5-Coder-3B trained together.}
    \label{fig:3b_exp_results}
\end{figure}

 We further test F-TIS in collaborative setups, where model size is the same, but models may have different expertise. We study two setups - collaborative training of Qwen2.5-1.5B and Qwen2.5-Coder-1.5B models, and of Qwen2.5-3B and Qwen2.5-Coder-3B models. The validation curves are reported in Figures \ref{fig:1b_exp_results} and \ref{fig:3b_exp_results}. 

\subsection{Trainable Parameters Heterogeneity}
\par
\begin{figure}[tb]
    \centering
    \begin{subfigure}{0.23\textwidth}
        \centering
        \includegraphics[width=\textwidth]{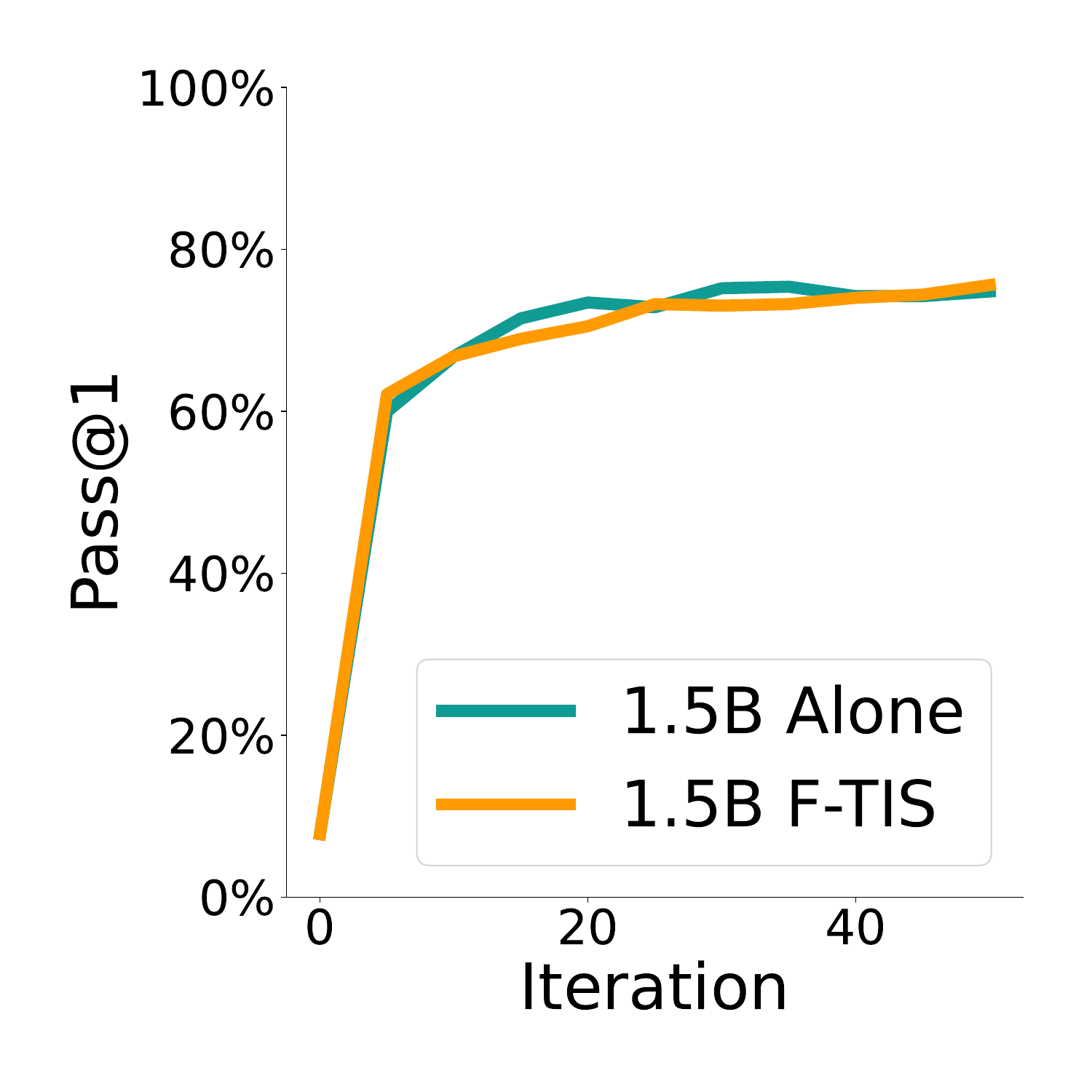}
        
        \caption{1.5B Model}

    \end{subfigure}
    \begin{subfigure}{0.23\textwidth}
        \centering
        \includegraphics[width=\textwidth]{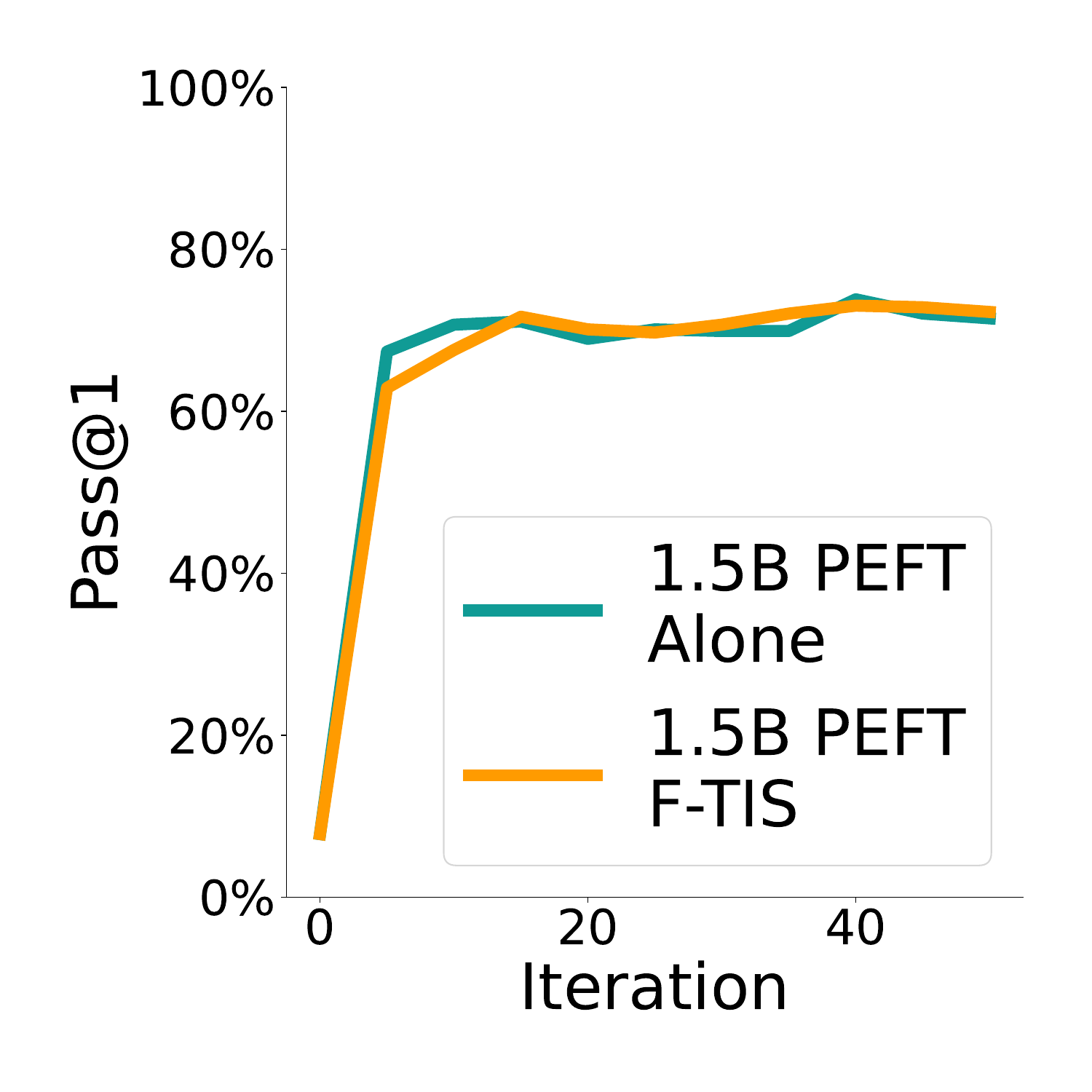}
        
        \caption{1.5B PEFT Model}

    \end{subfigure}
    \caption{Validation curves of a Qwen2.5-1.5B and a Qwen2.5-Coder-1.5B with LoRA trained together.}
    \label{fig:1b_peft_results}
\end{figure}

\begin{figure}[tb]
    \centering
    \begin{subfigure}{0.23\textwidth}
        \centering
        \includegraphics[width=\textwidth]{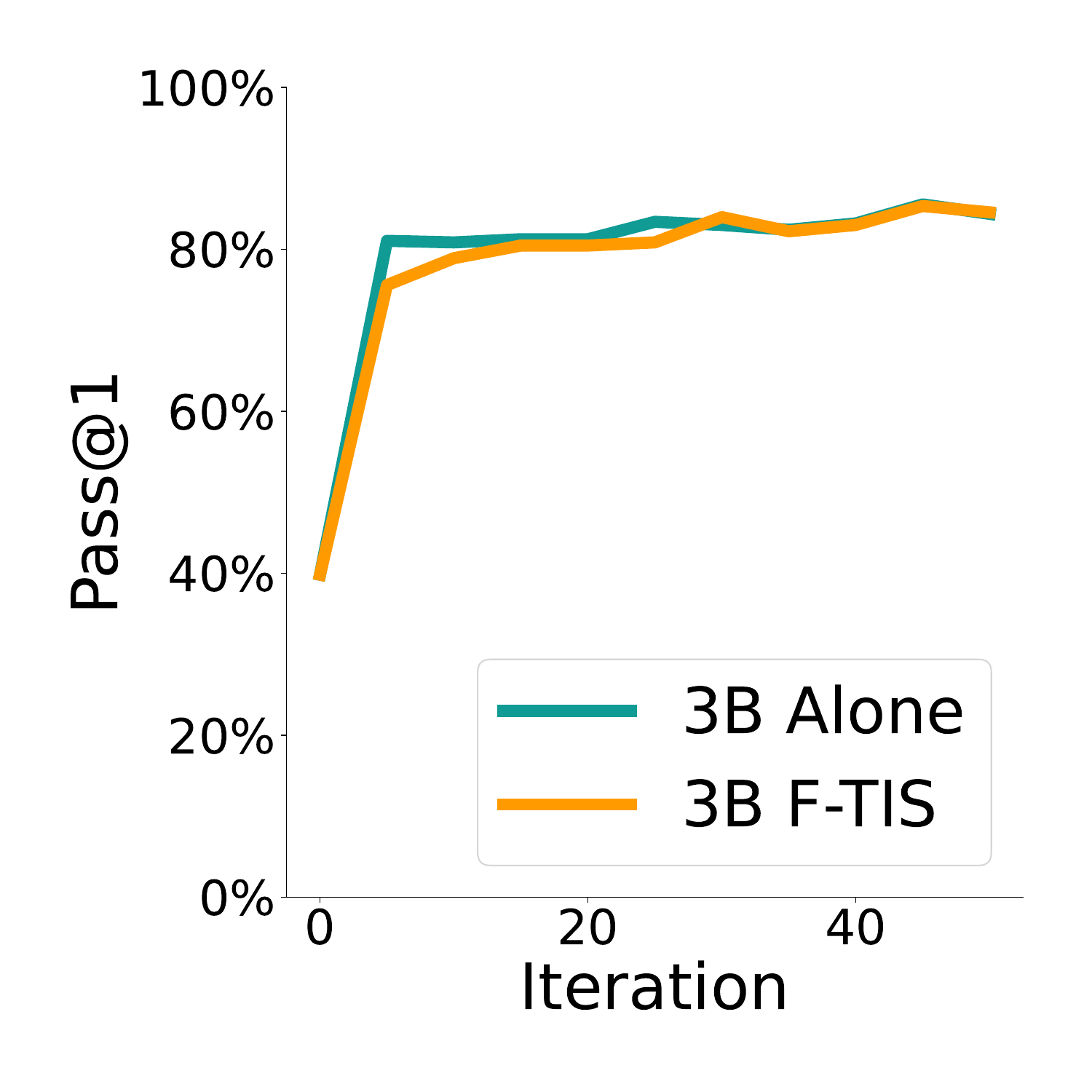}
        
        \caption{3B model}

    \end{subfigure}
    \begin{subfigure}{0.23\textwidth}
        \centering
        \includegraphics[width=\textwidth]{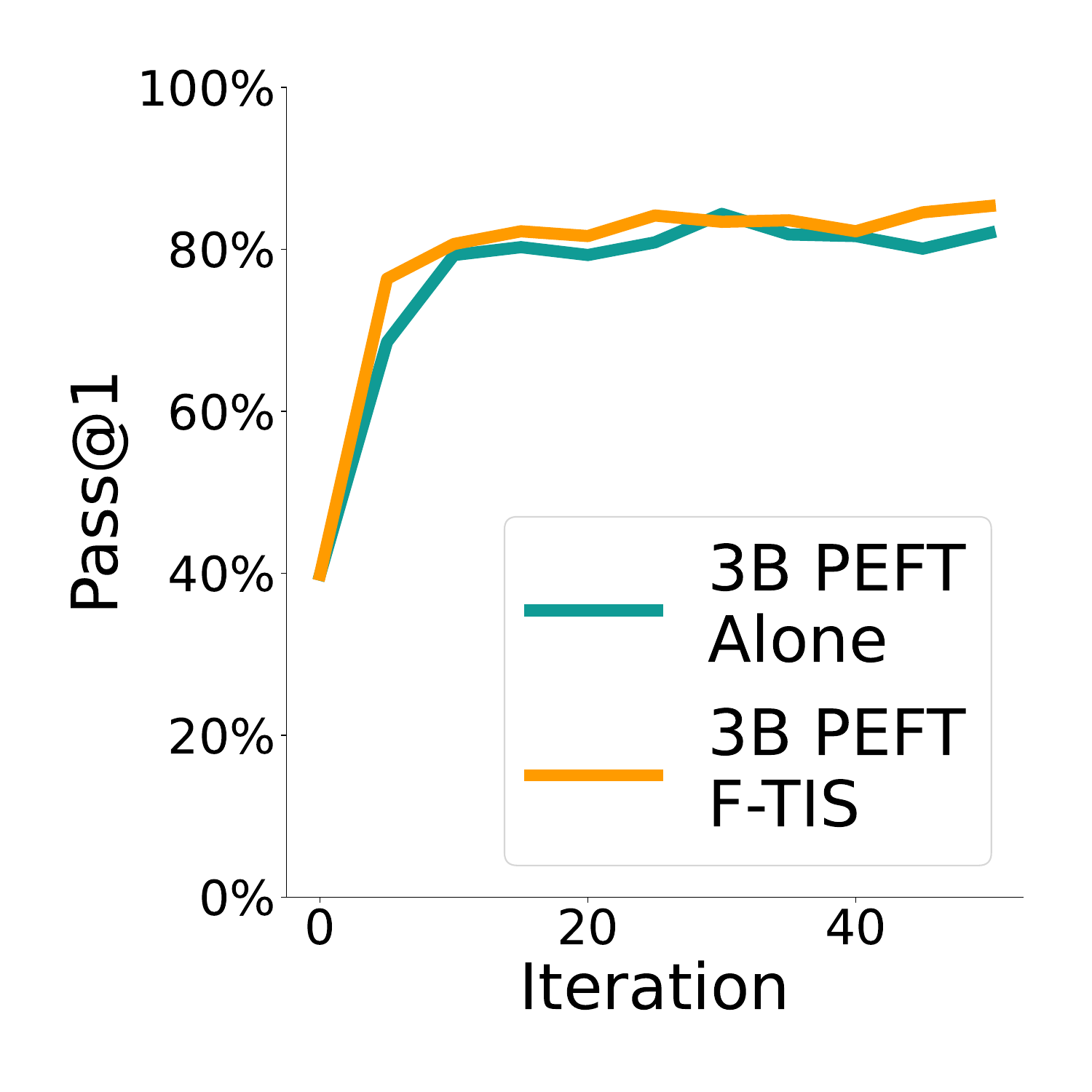}
        
        \caption{3B PEFT model}

    \end{subfigure}
    \caption{Validation curves of a Qwen2.5-3B and a Qwen2.5-3B with LoRA trained together}
    \label{fig:3b_peft_results}
\end{figure}

 To test the final model heterogeneity scenario, we perform collaborative GRPO with F-TIS on: Qwen-2.5-1.5B with and without LoRA, and for Qwen-2.5-3B with and without LoRA \cite{DBLP:journals/corr/abs-2106-09685}. We use LoRA with dimension of \(128\), targeting the Query and Value matrices. We present the validation curves in Figures \ref{fig:1b_peft_results} and \ref{fig:3b_peft_results}. Interestingly, we observe much better convergence of the 3B model with LoRA when trained collaboratively with the base model. This is further collaborated by the performance on out-of-distribution tasks in Table \ref{tab:math500}. This suggests that GRPO training with PEFT can be improved by using off-policy samples of a non-PEFT model.

\subsection{Out-of-distribution Math reasoning}

\par We evaluate all trained models on Math reasoning on the MATH-500 dataset \cite{huggingfaceh4_math500}, using the same system prompt and reward as before. This constitutes and out-of-distribution test - data that, albeit similar, has very different distribution. We present the results in the third column of Table \ref{tab:math500}, where models are grouped together with the respective model they were trained collaboratively with. Throughout all collaborations we observe a general trend where the model which performed worse off in the alone baseline can improve through the joint training. Most noticeably, the 3B PEFT model improves its performance by 7\%. However, the model which performed better in the baseline case, tends to perform worse off in the collaborative setting on out-of-distribution evaluation. A notable exception is the 3B Coder model, which when paired with a Base model improves its performance by 5.2\% and when paired with a smaller Coder model - by 12\%. We attribute this to a potential over-fitting of the larger Coder model to coding tasks, reducing their reasoning ability in Math tasks.

\subsection{Ablation on filtering}
\label{sec:g_ablation}

\par
\begin{figure}[tb]
    \centering
    \begin{subfigure}{0.23\textwidth}
        \centering
        \includegraphics[width=\textwidth]{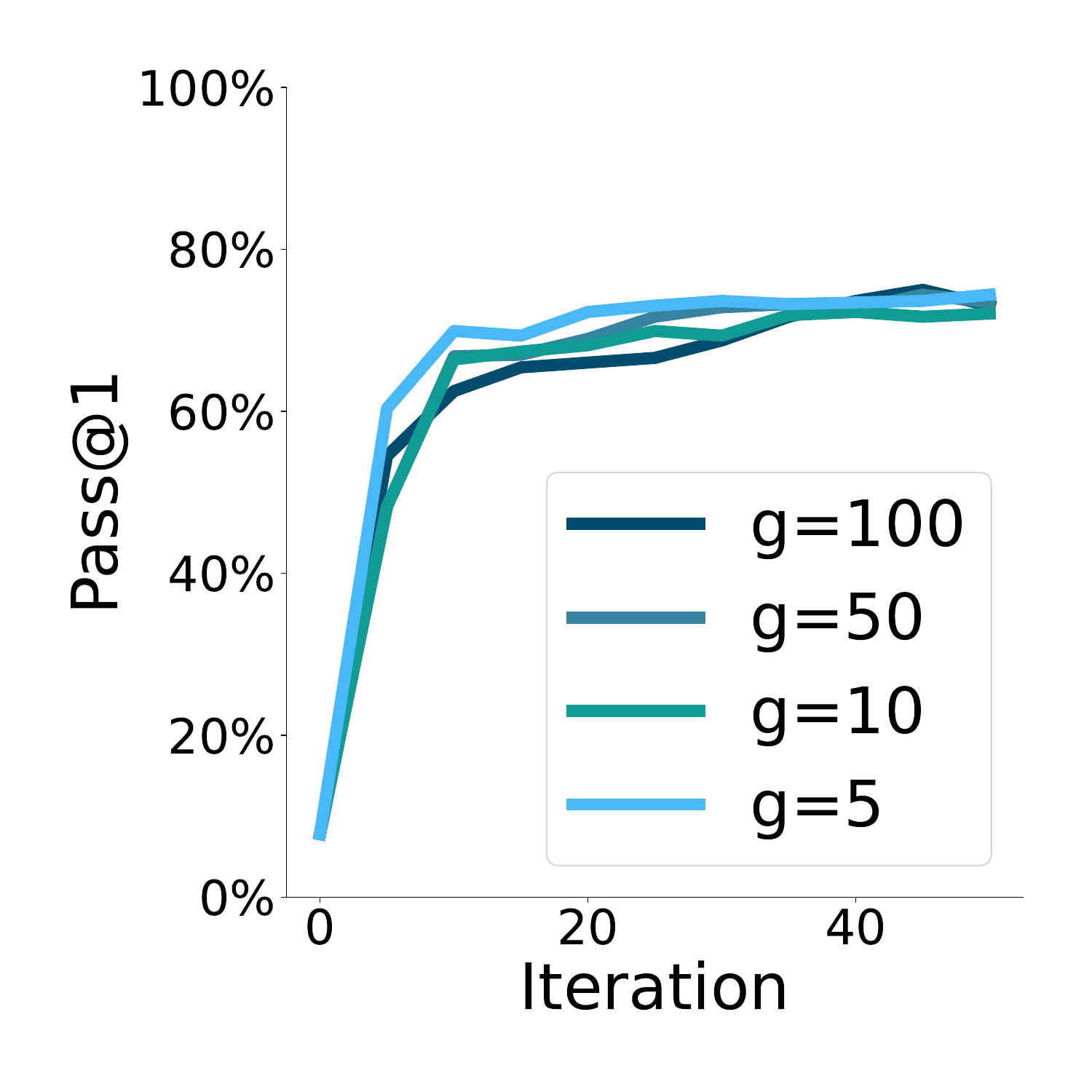}
        
        \caption{1.5B model}

    \end{subfigure}
    \begin{subfigure}{0.23\textwidth}
        \centering
        \includegraphics[width=\textwidth]{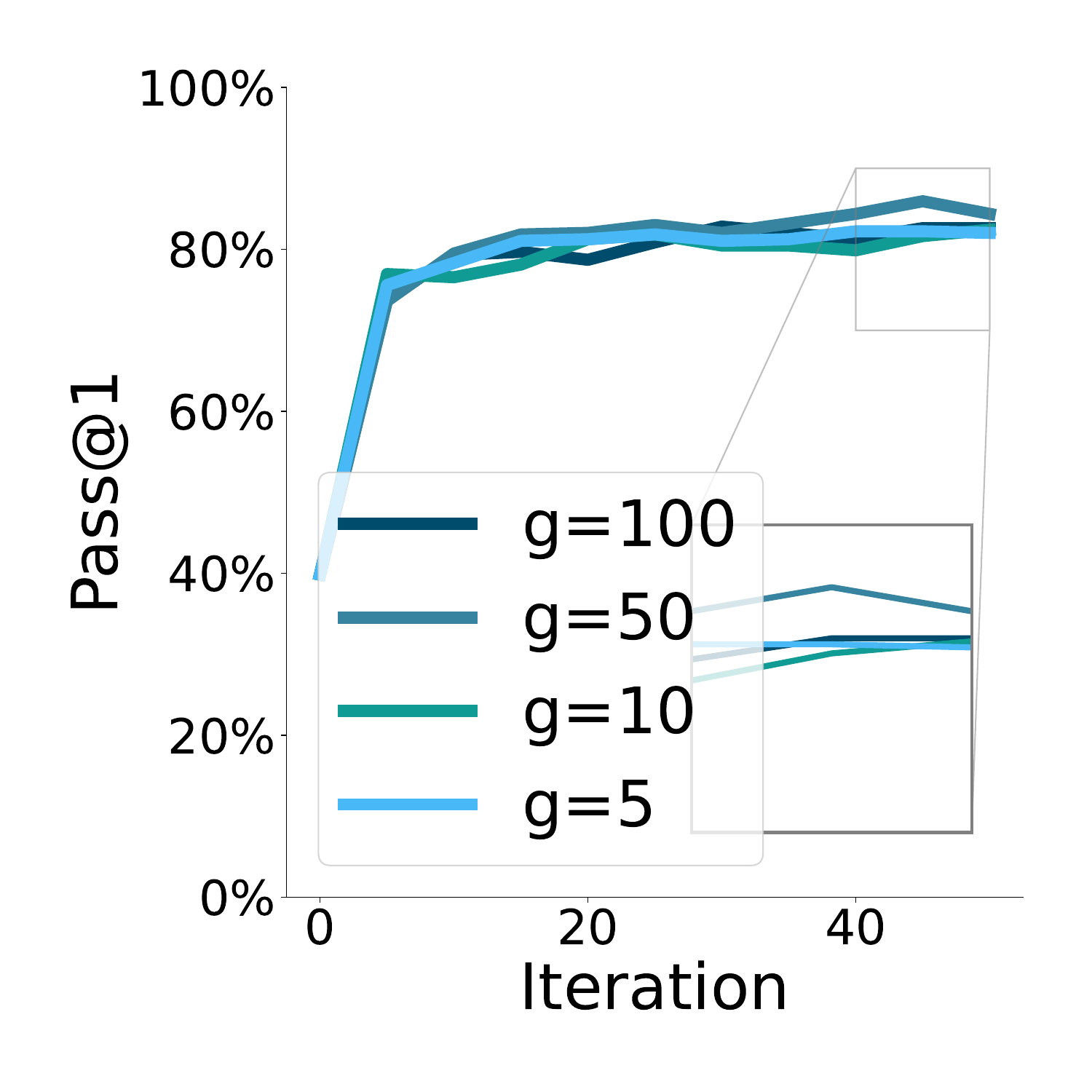}
        
        \caption{3B model}

    \end{subfigure}
    \caption{Ablation on the choice of \(g\) for F-TIS.}
    \label{fig:g_results}
\end{figure}

 Prior work has not studied suitable values for \(g\) in GRPO-training. Throughout these experiments we have solely used \(g=50\), as empirically we found it to be the best performant constant. In this section, we perform ablations to justify our choice of the value for this hyperparameter, repeating the experiments of Section \ref{sec:results_size} with \(g\) of 5, 10, 50, and 100. We report the validation curves in Figure \ref{fig:g_results}. Interestingly, we observe that for the 1.5B, especially initially, using a small \(g\) provides the greatest improvement in results. However, for the 3B model, using slightly higher \(g=50\) provides better performance. We attribute this to the fact that initially, models may produce really random outputs, while still learning the task. Such outputs provide the 1.5B with no good signal on how to improve, thus stalling its training. However, especially later in the training, the larger model benefits from the low-reward off-policy completions as it allows for greater exploration of the space. 

\subsection{F-TIS vs F-VIS}
\par
\begin{figure}[tb]
    \centering
    \begin{subfigure}{0.23\textwidth}
        \centering
        \includegraphics[width=\textwidth]{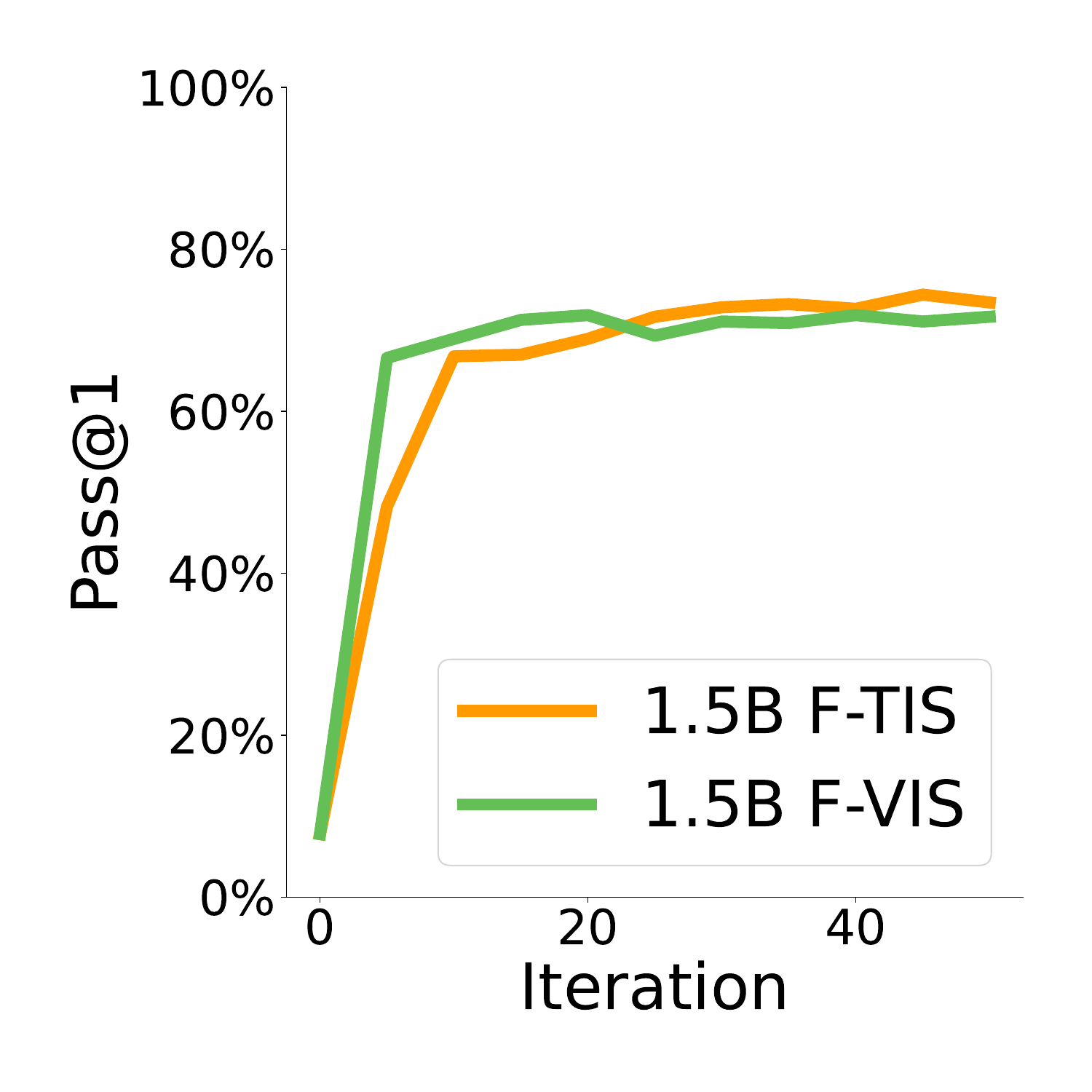}
        
        \caption{1.5B model}
        \label{fig:1Bfvis}

    \end{subfigure}
    \begin{subfigure}{0.23\textwidth}
        \centering
        \includegraphics[width=\textwidth]{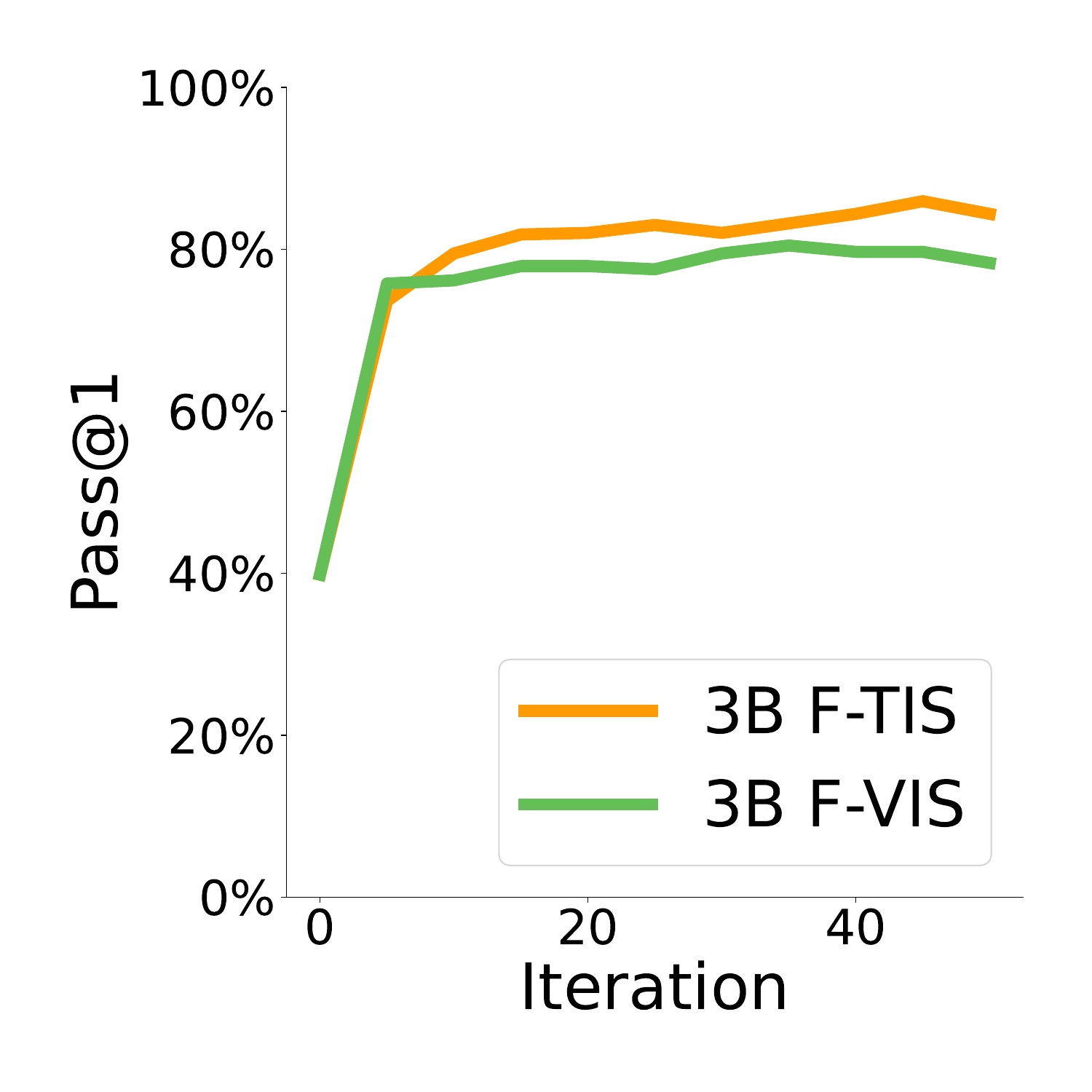}

        \caption{3B model}
        \label{fig:3Bfvis}

    \end{subfigure}
    \caption{Comparison between F-VIS and F-TIS.}
    
\end{figure}

 In our previous experiments we employed Truncated Importance Sampling as the base of our approach. Here we demonstrate that, even with filtering, Vanilla Importance Sampling underperforms compared to F-TIS. To this end we repeat the experiments of the size heterogeneity tests with F-VIS with \(g=50\) for both F-TIS and F-VIS. We report the validation curves of both approaches in Figure \ref{fig:1Bfvis} for the 1.5B model and Figure \ref{fig:3Bfvis} for the 3B model. We note the faster earlier convergence for F-VIS (similar to lower \(g\) in Figure \ref{fig:g_results}), however failing to generalize in later iterations, unlike F-TIS.

\subsection{Horizontal Collaboration}
\par
\begin{figure}[tb]
    \centering
    \begin{subfigure}{0.23\textwidth}
        \centering
        \includegraphics[width=\textwidth]{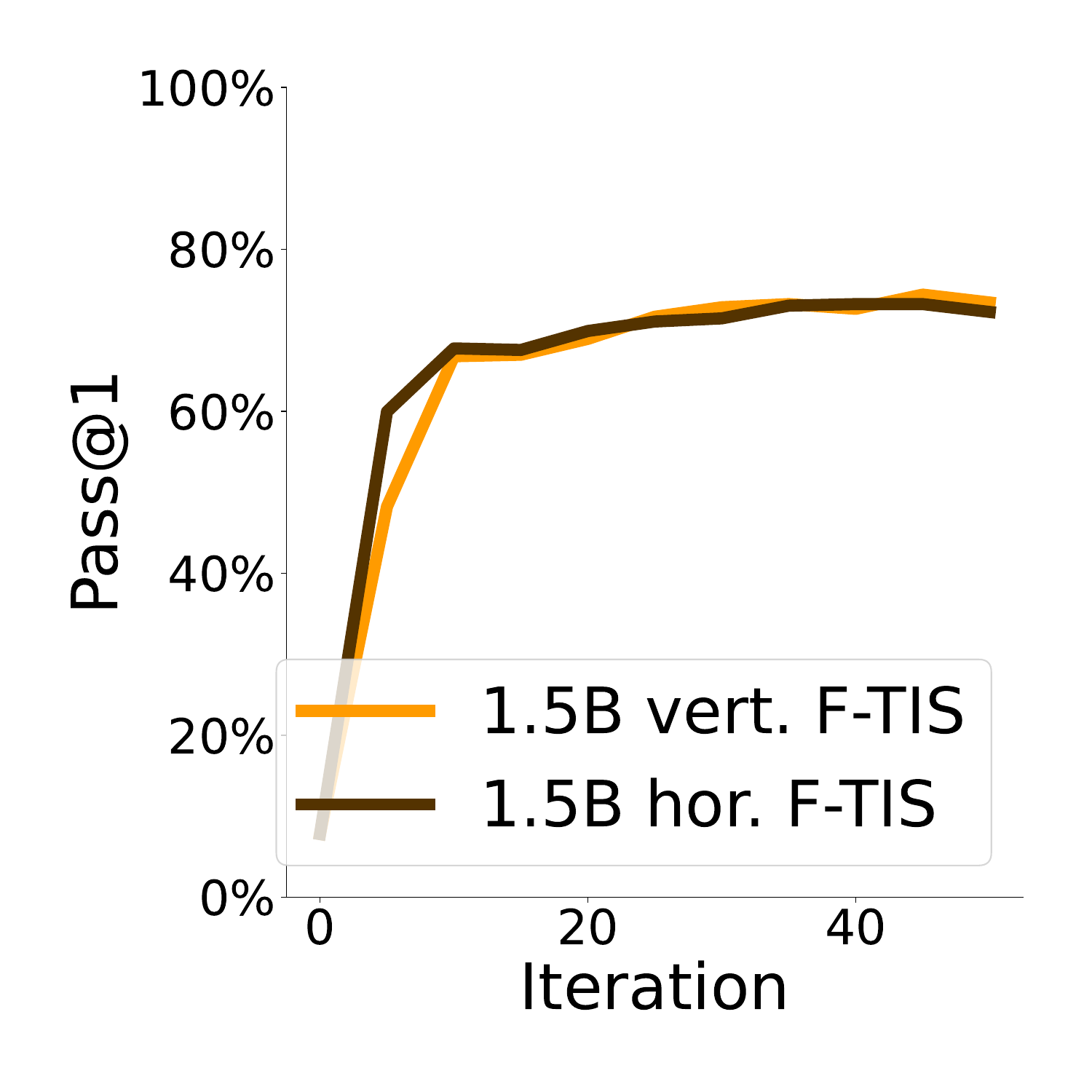}
        
        \caption{1.5B model}

    \end{subfigure}
    \begin{subfigure}{0.23\textwidth}
        \centering
        \includegraphics[width=\textwidth]{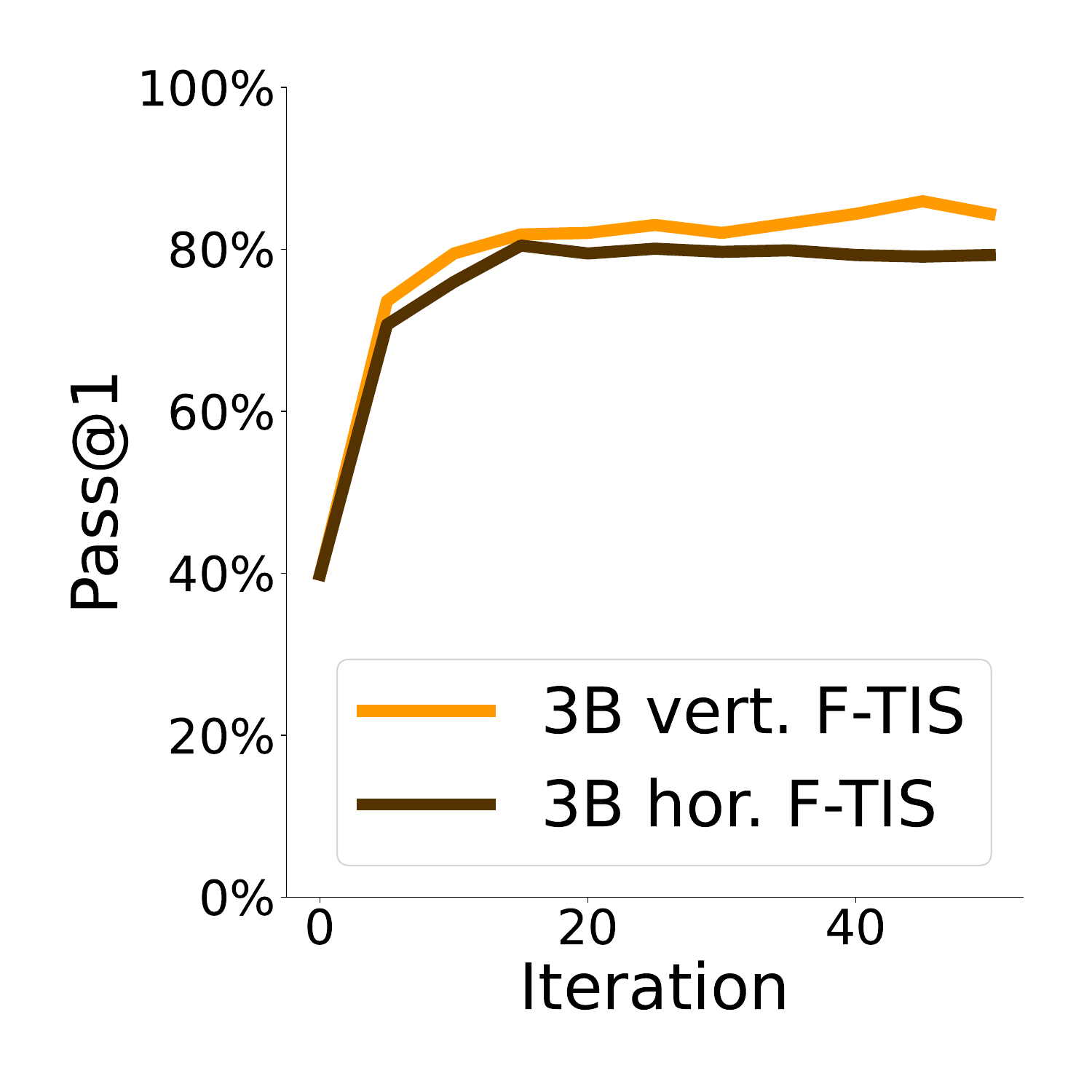}

        \caption{3B model}

    \end{subfigure}
    
    \caption{Comparison between horizontal and vertical F-TIS.}
    \label{fig:horizontal}
\end{figure}

 So far we have solely focused on vertical collaboration - one node responsible for all completions per question \cite{httt}. This has the benefit of faster generation (as nodes can parallelize computation independently for some subset of questions), requires less synchronization, and maintains the advantage calculation only with respect to one model's performance. We hypothesize that the last of these listed benefits would make horizontal learning impractical, as now the advantage calculation is performed relative to the swarm's mean performance. This could introduce an unwanted bias in the computation of the policy's gradient. We verify this hypothesis by repeating the experiments for F-TIS in Section \ref{sec:results_size}, however with horizontal collaboration. We present the results in Figure \ref{fig:horizontal}. As expected, horizontal learning leads to a noticeable degradation in the model's performance, though mainly noticeable for the 3B model.

\section{Conclusion}
\par In this paper, we presented F-TIS - a novel approach to collaborative training across different models in GRPO-style setups. We extensively evaluated our method in different heterogeneous settings, where model size, expertize, or trainable parameters can differ. For all, we observe similar convergence of F-TIS to that of fully on-policy learning. For some cases we even observe noticeable improvement of the final model on both in- and out-of-distribution tasks. We posit this work as one of the first works to study collaborative RL for LLMs and we hope it inspires further research into this new area.

\bibliography{ftis}
\bibliographystyle{plainnat}

\end{document}